\documentclass[pdflatex,sn-mathphys-num]{sn-jnl}% Math and Physical Sciences Numbered Reference Style 
\usepackage{graphicx}%
\usepackage{multirow}%
\usepackage{amsmath,amssymb,amsfonts}%
\usepackage{amsthm}%
\usepackage{mathrsfs}%
\usepackage[title]{appendix}%
\usepackage{xcolor}%
\usepackage{textcomp}%
\usepackage{manyfoot}%
\usepackage{booktabs}%
\usepackage{algorithm}%
\usepackage{algorithmicx}%
\usepackage{algpseudocode}%
\usepackage{listings}%
\usepackage{longtable}

\theoremstyle{thmstyleone}%
\theoremstyle{thmstyletwo}%

\theoremstyle{thmstylethree}%

\begin{document}

\title[The OxMat dataset: a multimodal resource for the development of AI-driven technologies in maternal and newborn child health]{The OxMat dataset: a multimodal resource for the development of AI-driven technologies in maternal and newborn child health}


% \author*[1,2]{\fnm{First} \sur{Author}}\email{iauthor@gmail.com}



% \affil*[1]{\orgdiv{Department}, \orgname{Organization}, \orgaddress{\street{Street}, \city{City}, \postcode{100190}, \state{State}, \country{Country}}}

% \affil[2]{\orgdiv{Department}, \orgname{Organization}, \orgaddress{\street{Street}, \city{City}, \postcode{10587}, \state{State}, \country{Country}}}

% \affil[3]{\orgdiv{Department}, \orgname{Organization}, \orgaddress{\street{Street}, \city{City}, \postcode{610101}, \state{State}, \country{Country}}}

\author{\fnm{M. Jaleed} \sur{Khan}}\email{jaleed.khan@wrh.ox.ac.uk}
\author{\fnm{Ioana} \sur{Duta}}\email{ioana.duta@wrh.ox.ac.uk}
\author{\fnm{Beth} \sur{Albert}}\email{beth.albert@wrh.ox.ac.uk}
\author{\fnm{William} \sur{Cooke}}\email{william.cooke@wrh.ox.ac.uk}
\author{\fnm{Manu} \sur{Vatish}}\email{manu.vatish@wrh.ox.ac.uk}
\author*{\fnm{Gabriel} \sur{Davis Jones}}\email{gabriel.jones@wrh.ox.ac.uk}
\affil{\orgdiv{Nuffield Department of Women's \& Reproductive Health (NDWRH)}, \orgname{University of Oxford}, \orgaddress{\street{Women's Centre, John Radcliffe Hospital}, \city{Oxford}, \postcode{OX3 9DU}, \country{United Kingdom}}}
%\affil[2]{\orgdiv{Department}, \orgname{University of Oxford}, \orgaddress{\street{}, \city{Oxford}, \postcode{OX3 9DU}, \country{United Kingdom}}}


\abstract{The rapid advancement of Artificial Intelligence (AI) in healthcare presents a unique opportunity for advancements in obstetric care, particularly through the analysis of cardiotocography (CTG) for fetal monitoring. However, the effectiveness of such technologies depends upon the availability of large, high-quality datasets that are suitable for machine learning. This paper introduces the Oxford Maternity (OxMat) dataset, the world's largest curated dataset of CTGs, featuring raw time series CTG data and extensive clinical data for both mothers and babies, which is ideally placed for machine learning. The OxMat dataset addresses the critical gap in women's health data by providing over 177,211 unique CTG recordings from 51,036 pregnancies, carefully curated and reviewed since 1991. The dataset also comprises over 200 antepartum, intrapartum and postpartum clinical variables, ensuring near-complete data for crucial outcomes such as stillbirth and acidaemia. While this dataset also covers the intrapartum stage, around 94\% of the constituent CTGS are antepartum. This allows for a unique focus on the underserved antepartum period, in which early detection of at-risk fetuses can significantly improve health outcomes. Our comprehensive review of existing datasets reveals the limitations of current datasets: primarily, their lack of sufficient volume, detailed clinical data and antepartum data. The OxMat dataset lays a foundation for future AI-driven prenatal care, offering a robust resource for developing and testing algorithms aimed at improving maternal and fetal health outcomes.}




\maketitle

% \section{Notes from GDJ}
% \begin{enumerate}
%     \item The best way to structure describing data of these kind for a clinically-focused data science audience is to describe them by the stage of pregnancy, e.g. Antepartum, Intrapartum, Postpartum, etc. and patient (e.g. mother or fetus/baby). Therefore, I would recommend describing the data by:
%     \begin{itemize} 
%             \item Pregnancy Stage (Maternal medical history prior to pregnancy, antepartum (the phase of pregnancy before labour), intrapartum (labour), postpartum (the time directly after labour), postnatal (0-24 days of life), etc.
%             \item Source Patient (e.g. Mother or baby)
%             \item Data/Information Source: E.g. electronic patient records, ultrasound, biochemistry, cardiotocography, etc.
%           \end{itemize}
%     \item Regarding the introduction, I would spend at least one paragraph describing the substantial data and AI gap in Women's health. Despite significant advances in the other medical domains (e.g. cardiology, radiology, etc.), Women's Health is under-represented. The community is becoming increasingly aware of this, e.g. https://www.weforum.org/publications/closing-the-women-s-health-gap-a-1-trillion-opportunity-to-improve-lives-and-economies/
% \end{enumerate}

\section{Introduction}\label{sec1} 

The growing application of Artificial Intelligence (AI) technologies in healthcare, estimated at a market size of \$22.45B in 2023 with a projected compound annual growth rate of 36.4\% through 2030 \cite{market2023}, highlights its potential and diverse applications across the sector. Significant investments, including Amazon's \$4B venture into Anthropic to advance generative AI \cite{amazon2023}, Corti's \$60M raise to reinforce real-time patient assessments \cite{corti2023}, Plilips' \$60M funding in AI-guided ultrasound for maternal health \cite{philips2023} and Aidoc's \$30M into medical imaging AI \cite{corti2023}, show the industry's confidence in leveraging AI for enhancing diagnostic accuracy, improving patient outcomes, and streamlining healthcare operations. Healthcare-targeted AI research marks a paradigm shift towards personalised medicine, enhancing patient care through improved diagnostics and treatment outcomes \cite{alowais2023revolutionizing}. AI-driven biomedical health monitoring systems using non-invasive sensors demonstrate the potential for early disease detection and accurate diagnosis \cite{sonawani2023biomedical}. Leveraging extensive datasets from wearable sensors and medical imaging, Machine Learning (ML) is significantly advancing precision medicine \cite{qureshi2023artificial}. The development of the Internet of Medical Things (IoMT) and point-of-care devices underscores the role of AI in facilitating remote health monitoring and diagnostics for various applications, including cardiac care and diabetes and cancer management \cite{manickam2022artificial}. The adoption of AI and wearable sensors is shifting healthcare delivery towards more patient-centred models by enabling real-time analysis of vital health parameters \cite{junaid2022recent}. The application of AI in radiology has demonstrated its potential in interpreting medical images with variability ranges comparable to those of experienced specialists \cite{choy2018current, rodriguez2019stand}. In addition, the application of AI in ultrasound examination and diagnosis has also shown promising results \cite{stirnemann2023development}. In the context of biomedical time series analysis, accurately interpreting time series signals (e.g. electrocardiograms, electroencephalograms, cardiotocograms) is vital, given their role in health monitoring and predictive analytics. The challenges of handling high-dimensional healthcare data and the critical need for feature engineering to capture temporal patterns are addressed through advanced deep learning techniques. These methods have demonstrated superior performance in understanding the complex nonlinear interactions within biomedical data \cite{morid2023time}. Advancements in signal processing techniques have further empowered development of intelligent systems for health monitoring, emphasising the potential and necessity of novel approaches in non-invasive biomedical health monitoring, disease detection and diagnosis \cite{sonawani2023biomedical}. Collectively, these developments underscore the transformative impact of AI and deep learning in interpreting biomedical image and time series data for more precise, efficient, and tailored healthcare to individual patient needs. \par

Despite the aforementioned significant advances in other medical domains, the World Economic Forum's report, in collaboration with the McKinsey Health Institute, reveals a proclaimed data and AI gap in women's health \cite{wec2024}. This gap, affecting areas from fundamental research to specific health condition advances, highlights an urgent need for more attention towards women's health, including reproductive and maternal-fetal care. Notably, women experience 25\% more of their lifespan in poor health than men, indicating a critical area for improvement. This disparity is attributed to insufficient research investment and the underestimation of women's health issues. Addressing this gap holds substantial economic potential, with projections suggesting it could contribute \$1 trillion annually to the global economy by 2040 \cite{wec2024}. Within the domain of obstetric healthcare, cardiotocography (CTG), also referred to as electronic fetal monitoring, stands as a ubiquitous principal method for fetal surveillance in prenatal care. CTG uses ultrasound to measure the fetal heart rate (FHR) and uterine activity alongside their temporal relationships to facilitate the monitoring of the fetal state before and during labour. The Dawes-Redman (DR) computerized CTG system \cite{pardey2002computer}, a pioneer in identifying CTG patterns reflecting fetal sleep states, established the gold standard for computerised CTG analysis in clinical practice. CTG guides clinical decision making to mitigate risks of adverse neonatal outcomes, such as neonatal acidaemia, hypoxia, stillbirth, and cerebral palsy \cite{ayres2015figo}. Despite its widespread use and the pivotal role it plays in obstetrics, the efficacy of CTG in reducing adverse neonatal outcomes has been a subject of debate, primarily due to the inherent intra- and inter-observer variability in signal interpretation and its low specificity. Since the introduction of CTG in the 1960s, there have been reports of an increase in cesarean section rates and instrumental vaginal births without a definitive reduction in the incidence of adverse neonatal outcomes \cite{alfirevic2017continuous}. Accurately interpreting CTG signals is essential for detecting potential complications and guiding timely interventions to prevent fetal harm, as well as identifying healthy pregnancies to avoid unnecessary interventions.\cite{alyousif2021systematic}. However, the quality of CTG interpretation is significantly influenced by the observer's training and experience, with factors such as prolonged working hours, heavy workloads, and limited exposure to atypical cases affecting the consistency and reliability of interpretations \cite{aeberhard2024introducing}. These challenges underscore the limitations of manual CTG analysis and highlight the need for automated CTG analysis for enhanced objectivity and reproducibility in fetal monitoring. The integration of AI and ML in CTG analysis significantly advances the precision and reliability of prenatal monitoring. Given the challenges posed by the non-stationarity and interference in FHR signals, AI and ML offer effective solutions for noise reduction, feature detection, fetal state classification and impending adverse outcome detection, showcasing a significant improvement over traditional methods \cite{aeberhard2023artificial}. These technologies excel in addressing the issues of inter- and intra-observer variability, ensuring a more objective and consistent assessment of fetal well-being. Moreover, the potential of AI and ML to become integral to clinical practice lies in their ability to standardize CTG interpretation and eliminate biases \cite{barnova2024artificial}. \par

The core scientific principle behind these technologies and advances revolves around ML-based analysis of physiological time series data. Preliminary research into CTG analysis exists, yet it has been constrained by limitations in dataset size and standard pre-processing and evaluation protocols. The literature contains several CTG datasets \cite{aeberhard2024introducing, brocklehurst2017computerised, chudacek2014open, georgieva2019computer, uci2024ctg}, but most lack sufficient volume and detailed clinical, neonatal, and postnatal information, primarily focusing on the intrapartum stage. Notably, the largest dataset \cite{brocklehurst2017computerised, brocklehurst2017infant} comprises around 46,000 pregnancies, yet it concentrates on the intrapartum phase, neglecting the critical antepartum period where the majority of serious adverse outcomes, such as 80\% of stillbirths, could be preemptively identified \cite{hug2022global}. For comprehensive experimentation and selection of the optimal ML algorithms for CTG analysis tasks like risk stratification and early condition detection, a large, high-quality dataset encompassing a substantial volume of real physiological or pathological recordings is essential \cite{barnova2024artificial}. Such a dataset should include comprehensive clinical, neonatal, and postnatal outcomes, annotations, sufficient signal lengths, and data from fetuses at various gestational ages. This paper undertakes an exhaustive analysis of existing datasets (Section \ref{sec2}) and presents the collection (Section \ref{sec3}) and cleaning (Section \ref{sec4}) of our new OxMat dataset, the largest curated collection of CTGs with associated clinical data (Section \ref{sec5}) for both mothers and babies, featuring raw time series CTG data ideally placed for machine learning. Since 1991, all related pregnancy and CTG data have been meticulously collected, with each data point undergoing weekly quality control by clinical experts. Unlike many existing datasets where CTGs are recorded in paper format, limiting their utility for machine learning, the OxMat dataset includes raw, digital time series CTG data. The OxMat dataset currently includes 177,211 unique CTGs from 51,036 pregnancies, with an annual addition of 5,711 CTGs from 1,646 pregnancies on average from 1991 to 2021. This data predominantly consists of 96.46\% singleton births and 94.23\% antepartum CTGs. For each data point, over 200 clinical variables are documented, achieving nearly 100\% data completeness for critical outcomes like stillbirths or post-delivery fatalities. \par

\section{Existing Datasets}\label{sec2}

In this section, we comprehensively review and summarize the existing CTG datasets used in fetal monitoring research that employ signal processing and machine learning techniques. The key characteristics of each dataset, including the pregnancy stage, source, inclusion and exclusion criteria, volume and split, are presented in Table \ref{tab_datasets}. Our search spanned several online repositories, including Physionet, the UCI Machine Learning Repository, and Google Scholar, to curate a comprehensive list of related datasets available in the literature. \par

Previous work \cite{georgieva2013artificial} sourced a dataset from the John Radcliffe Hospital, Oxford, covering 107,614 deliveries between April 20, 1993, and February 28, 2008. Their final dataset comprised 7,568 data points with clinical data and CTG recordings. The data of 124 cases was used for training and 252 cases for testing, both with 50\% adverse outcomes and 50\% normal outcomes. The authors used feed-forward Artificial Neural Networks (ANN) to analyze Fetal Heart Rate (FHR) for predicting adverse labour outcomes. An ensemble of ten ANNs was trained on a dataset of 124 patients (a balanced set of 62 adverse and 62 normal outcomes). A total of 12 features, 6 from CTG and 6 from clinical data, were extracted. Principal Component Analysis (PCA) was used in the preprocessing to reduce these 12 dimensions to 6. The complete unbalanced dataset with 7,568 cases was used to assess the general applicability of the ANN models. The ANN ensemble achieved a sensitivity of 60.3\% and a specificity of 67.5\%, with a misclassification rate of 36\%. \par 

Petrozziello et al. \cite{petrozziello2019multimodal} collected a dataset including 35,429 births from the John Radcliffe Hospital, Oxford, covering data from 1993 to 2011, split into 85\% for training and 15\% for testing. Inclusions were based on CTG recordings from labour at 36 weeks of gestation or more, ending within three hours of birth, and with validated cord blood gas analysis. Exclusions were for breech presentation and congenital abnormalities. External validation was performed using datasets from the SPaM Workshop 2017 database \cite{georgieva2019computer} and the CTU-UHB database \cite{chudacek2014open}, including 200 and 552 cases, respectively. The authors employed multimodal deep learning to analyze CTG data for predicting fetal distress during labour. Their architecture comprises a stack of two Multimodal Convolutional Neural Networks (MCNN) to analyze CTG data in two stages of labour, using the output of the first stage as an additional input for the second stage. An MCNN processes the input signals, including fetal heart rate, uterine contractions, and signal quality scores, independently via two Convolutional Neural Networks (CNN) and a Fully Connected Network (FCN). The features from these networks are then integrated and processed by another FCN, followed by softmax classification.  A True Positive Rate (TPR) of 53\% with a fixed False Positive Rate (FPR) of 15\% is reported on the Oxford testing set. The authors acknowledged the influence of CTG signal quality on model performance and the current limitation of the models in detecting severe fetal injury without cord acidemia, suggesting a hybrid future approach integrating clinical knowledge and data-driven analysis for better CTG interpretation. \par

% \begin{table}[h!]
% \caption{Existing Datasets}
% \label{tab_datasets}
% \begin{tabular}{|l|p{1.4cm}|p{1.6cm}|p{3cm}|p{2cm}|p{1.6cm}|}
\begin{longtable}{|p{1.6cm}|p{1.8cm}|p{2.2cm}|p{3.2cm}|p{2.4cm}|p{1.6cm}|}
\hline %\toprule 
Publication & Pregnancy Stage & Data Source & Inclusion and Exclusion Criteria & Size \& Split & External Valid. Data \\
\hline %\midrule
Georgieva et al. \cite{georgieva2013artificial} & Intrapartum (2nd stage of labour) & 107,614 deliveries at John Radcliffe Hospital, Oxford (1993-2008) & Cases with incomplete labour data, non-labour Caesarean, multiple pregnancies, antepartum stillbirth, insufficient CTG signal quality, intrapartum stillbirth, prematurity, etc. excluded & 124 training, 252 testing and 7,568 total cases & --  \\ \hline %custom
Petrozziello et al. \cite{petrozziello2019multimodal} & Intrapartum (labour) & CTGs and clinical data from 35,429 births at John Radcliffe Hospital, Oxford & Deliveries at 36 weeks of gestation or more, CTGs longer than 15 minutes ending within 3 hours of birth, and validated cord blood gas analysis immediately after birth included; cases with breech presentation and congenital abnormalities excluded & 35,429 data points, categorized into 5 groups based on labour outcomes and cord arterial pH values, 85-15 train-test split & 300 cases from the SPaM 2017 workshop \cite{georgieva2019computer} (Lyon and Brno) and 552 cases from the CTU-UHB database \cite{chudacek2014open} \\ \hline %custom
Aeberhard et al. \cite{aeberhard2024introducing} & Intrapartum (labour) & 19,399 raw CTG recordings and outcomes at University Hospital of Bern (2006-2019) & Exclusions made for incomplete data, lack of consent and multiple births; fetal $pH \geq 7.15$ and 5-min $APGAR \geq 7$ considered physiological & 6,141 complete data points  & -- \\ \hline %custom 
INFANT \cite{brocklehurst2017computerised, brocklehurst2017infant} & Intrapartum & 47,062 women across 24 maternity units in the UK and Ireland (2010-2013) & Women in labour, aged 16+, undergoing continuous CTG monitoring, with singleton or twin pregnancies at 35+ weeks gestation included; women with known gross fetal abnormalities excluded & 46,042 women in the final dataset & -- \\ \hline %custom
CTU-UHB \cite{chudacek2014open} & Intrapartum & 9,164 CTGs recorded at University Hospital Brno (2010--2012) & CTGs lasting up to 90 minutes and starting a maximum of 90 minutes before delivery, mature fetuses ($\geq37$ weeks of gestation), singleton pregnancies and cases without known fetal diseases included; recordings with incomplete clinical data or those not meeting technical quality standards excluded & 552 CTGs with clinical and technical parameters, including the type of delivery (vaginal/cesarean section), maternal age, gestational age and outcomes such as umbilical artery pH and APGAR scores & -- \\ \hline %custom/standard
SPaM-2017 \cite{georgieva2019computer} & Intrapartum & 300 labour cases from centres in Lyon, Brno, and Oxford, including singleton pregnancies over 36 weeks of gestation & Singleton pregnancies $>36$ weeks gestation with validated cord gases included; small-for-gestational-age babies and those with congenital abnormalities excluded; CTGs selected based on length ($>60$ minutes), minimal signal loss ($<15\%$), and proximity to birth (ending within 10 min of delivery) & 300 & -- \\ \hline %custom
Feng et al. \cite{feng2023hybrid} & Intrapartum & 23,500 cases, including FHR signals, UC signals, and clinical data, of pregnant women with gestational ages of 28-42 weeks, collected at the affiliated hospitals of Guangzhou University of Chinese Medicine and Jinan University, China (2016--2018) & Inclusion based on obstetrician interpretation to assess suitability & 16,355 cases (11,998 normal and 4,357 abnormal) & -- \\ \hline
Signorini et al. \cite{signorini2016advanced} & Intrapartum & Data of 122 subjects related to normal and pathological fetuses collected from various Italian university clinics including Azienza Ospedaliera Universitaria Federico II, Naples & -- & 122 CTGs (61 normal and 61 IUGR) \& -- & -- \\ \hline
Lin et al. \cite{lin2024deep} & Antepartum & CTG recordings of 86 singleton deliveries at Peking University Third Hospital (2014--2018) & Singleton pregnancies with monitoring conducted between $28$ to $41^{+6}$ weeks of gestation included; fetal malformations and cases with $>50\%$ signal loss excluded & 114 long-term FHR sequences, 91--23 split & -- \\ \hline %custom 
Davis Jones et al. \cite{jones2024performance} & Antepartum & 4,196 term antepartum FHR traces extracted from Oxford University Hospitals database (1991--2021) & Term singleton pregnancies (37--42 gestational weeks) with complete clinical outcome data included; traces with significant data gaps or incomplete analyses excluded; for adverse outcomes, only traces performed within 48h before delivery considered & 4,196 FHR traces matched by fetal sex and gestational age, focusing on a healthy cohort versus an adverse outcome cohort & -- \\ \hline %custom
UCI \cite{uci2024ctg} & -- & University of California Irvine & -- & 2,126 CTGs automatically processed and classified by 3 obstetricians into 10 morphologic patterns and 3 fetal states & -- \\ \hline %custom/standard
% Paper & Stage & Source & Inc/Exc Criteria & Size \& Split & External Valid. Data \\
\caption{Existing Datasets}
\label{tab_datasets}
\end{longtable}
%\end{tabular}
%\end{table}

Aeberhard et al. \cite{aeberhard2024introducing} collected raw intrapartum CTG recordings from labouring women at the University Hospital of Bern between 2006 and 2019, subsequently matching these with corresponding fetal outcomes, such as arterial umbilical cord pH and 5-min APGAR score. An initial set of 19,399 CTG recordings was retrieved using the IntelliSpace Perinatal system. To address challenges like spelling discrepancies and missing data, the search process was refined to use Patient Identification Numbers (PIDs) for better accuracy. The cases lacking consent, having incomplete data and multiple births were systematically excluded. Physiological fetal pH was defined as 7.15 and above, and a 5-min APGAR score of $\geq 7$ was considered normal. Out of the 19,399 CTG recordings, clinical outcomes were initially found for 3,400 cases. Following a refined search strategy, 6,141 complete data samples were obtained. The final dataset comprising 6,141 paired data samples, including both CTG raw data and corresponding maternal and fetal outcomes, represents 31.65\% of the initial CTG recordings. The study aimed to develop predictive models for fetal hypoxia. Half of the dataset is planned to be used to train AI models, while the other half is planned to be used for efficacy analysis. The paper highlighted the potential for AI-assisted CTG interpretation to support clinical decision-making during delivery while acknowledging the inherent challenges in the process of data curation and the creation of large datasets with patient outcomes from different sources. \par

The INFANT Collaborative Group at the University of Oxford created the currently largest database and explored the efficacy of decision support software on CTG interpretations and neonatal outcomes during labour \cite{brocklehurst2017computerised, brocklehurst2017infant}. Data were collected from 47,062 women at 24 maternity units across the UK and Ireland, from January 6, 2010, to August 31, 2013, focusing on women undergoing continuous electronic fetal monitoring. This dataset was specifically curated for this study, utilizing the INFANT system, a decision support software developed to enhance CTG interpretation alongside other clinical information. Eligibility for participation included women in labour, aged 16 or older, with singleton or twin pregnancies at 35 weeks gestation or more, excluding those with known gross fetal abnormalities. Following exclusions for incomplete consent forms and withdrawals, the final analysis comprised data from 46,042 women. The trial aimed to determine if computerized assistance in interpreting CTGs could reduce adverse neonatal outcomes. However, the results revealed no significant difference between the groups, with both showing a 0.7\% incidence of poor outcomes and an adjusted risk ratio of 1.01 (95\% CI 0.82–-1.25). \par

The CTU-CHB database \cite{chudacek2014open}, available on Physionet, includes 552 CTG recordings from vaginal deliveries and cesarean sections, selected from over 9,164 recordings collected at the University Hospital in Brno, Czech Republic between April 2010 and August 2012. The selection criteria focused on mature fetuses (37 weeks of gestation or more) and singleton pregnancies, excluding known fetal diseases or conditions that could bias FHR data. Each recording in the final database lasts up to 90 minutes and is initiated no more than 90 minutes before delivery. Romagnoli et al. \cite{romagnoli2020annotation} enhanced the existing CTU-CHB database \cite{chudacek2014open} from Physionet with expert annotations for 552 CTG recordings captured during the intrapartum stage of labour. This dataset includes annotations for the FHR and maternal tocogram signals, including specific CTG events like bradycardias, tachycardias, accelerations, decelerations and uterine contractions, for automating CTG analysis. An expert gynaecologist performed the annotation using CTG Analyzer \cite{sbrollini2017ctganalyzer}, an automated CTG analysis software. \par

Feng et al. \cite{feng2023hybrid} used the UCI dataset \cite{uci2024ctg} and data of 23,500 fetal monitoring cases for pregnancies within 28-42 weeks gestational ages collected at the affiliated hospitals of Guangzhou University of Chinese Medicine and Jinan University in China between 2016 and 2018 to evaluate ML techniques for enhancing the accuracy and interpretability of CTG classification. Based on obstetrician interpretation to assess suitability for inclusion and classification, 16,355 cases were selected. A stacked ensemble strategy was employed, leveraging Support Vector Machines (SVM), eXtreme Gradient Boosting (XGB) and Random Forests (RF) as base learners, with backpropagation neural networks serving as the meta-learner. The authors reported 0.9539 accuracy and a 0.9249 average F1 score on the UCI dataset and 0.9201 accuracy with a 0.8926 average F1 score on the private dataset. \par 

Signorini et al. \cite{signorini2016advanced} collected and analysed FHR signals collected from 122 cases during the intrapartum stage to assess fetal well-being during labour using time-domain and frequency-domain signal processing techniques. The data was collected at various Italian university clinics, with a significant portion coming from the Azienza Ospedaliera Universitaria Federico II in Naples, comprising recordings from the gestational age range of 24 to 42 weeks, including both normal and pathological cases, with a particular focus on those diagnosed with intrauterine growth restriction (IUGR). \par

Lin et al. \cite{lin2024deep} curated a dataset from Peking University Third Hospital, comprising 114 sequences from 86 singleton deliveries between April 2014 and December 2018. Inclusion criteria targeted singleton pregnancies with adequate monitoring data, excluding cases with fetal malformations or significant data loss. The authors introduced the Long-term Antepartum Risk Analysis (LARA) system for automated analysis of long-term FHR data using deep learning and information fusion. It utilizes a CNN alongside information fusion methods to process the FHR data, producing a Risk Distribution Map (RDM) and a Risk Index (RI) to evaluate fetal health risks. The performance evaluation showed an AUC of 0.872, accuracy of 0.816, specificity of 0.811, sensitivity of 0.806, precision of 0.271, and F1 score of 0.415. The study reported a significant correlation between higher RI and adverse outcomes (p=0.0021). \par

Jones et al. \cite{jones2024performance} evaluated the effectiveness of the Dawes-Redman computerized Cardiotocography (DR-cCTG) algorithm in assessing antepartum fetal well-being. The Dawes-Redman (DR) system's criteria \cite{pardey2002computer} for normality is given in Table \ref{tab:dr_criteria}. The study used 4,196 antepartum FHR recordings from term singleton pregnancies, extracted from the Oxford University Hospitals database between 1991 and 2021. The traces with complete clinical outcomes, excluding those with significant data gaps or incomplete Dawes-Redman analyses, were considered. For performance evaluation against adverse outcomes, only FHR traces conducted within 48 hours prior to delivery were considered. The performance was evaluated using accuracy, sensitivity, specificity, and predictive values in various scenarios, including different risk prevalences and temporal proximity to delivery. The key findings include a high sensitivity (91.7\%) for detecting fetal well-being but low specificity (15.6\%) for adverse outcomes, with predictive values varying based on population prevalence. The study concludes that while DR-cCTG is effective in detecting fetal well-being, especially in low-risk settings, its utility in high-risk pregnancy scenarios and for predicting adverse outcomes is limited. \par

\begin{table}[h]
\caption{The Dawes-Redman (DR) system's criteria for normality (reproduced from  \cite{pardey2002computer})}\label{tab:dr_criteria}
\begin{tabular}{|p{12cm}|}
\hline
1. The recording must contain at least one episode of high variation. \\
2. The STV must be $>$3.0 ms, but if it is $<$4.5 ms the LTV averaged across all episodes of high variation must be $>$3rd percentile for gestational age. \\
3. There must be no evidence of a high-frequency sinusoidal rhythm. \\
4. There must be at least one acceleration, or a fetal movement rate $\geq$20 per hour and an LTV averaged across all episodes of high variation that is $>$10th percentile for gestational age. \\
5. There must be at least one fetal movement or three accelerations. \\
6. There must be no decelerations $>$20 lost beats if the recording is $<$30 minutes, no more than one deceleration of 21–-100 lost beats if it is $>$30 minutes, and no decelerations at all $>$100 lost beats. \\
7. The basal heart rate must be 116-–160 beats/min if the recording is $<$30 minutes. \\
8. The LTV must be within 3 standard deviations of its estimated value or (a) the STV must be $>$5.0 ms, (b) there must be an episode of high variation with $\geq$ 0.5 fetal movements per minute, (c) the basal heart rate must be $\geq$ 120 beats/min, and (d) the signal loss must be $<$30\%. \\
9. The final epoch of the recording must not be part of a deceleration if the recording is $<$60 minutes or a deceleration at 60 minutes must not be $>$20 lost beats. \\
10. There must be no suspected artefacts at the end of the recording if the recording is $<$60 minutes. \\
\hline
\end{tabular}
\end{table}

\section{Data Collection and Consolidation}\label{sec3}

\begin{figure}[ht!]
\centering
\includegraphics[width=\textwidth]{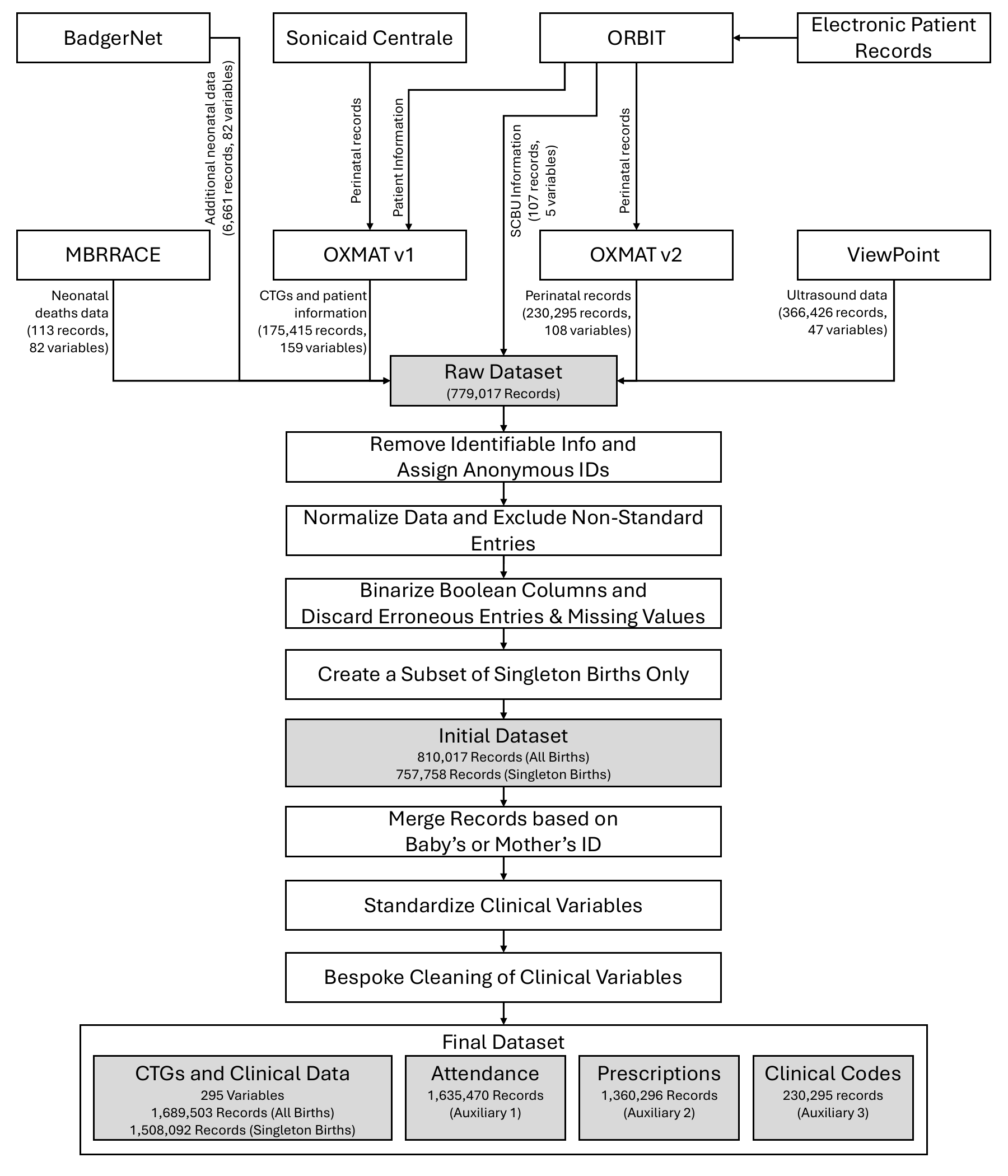}
\caption{Dataset Curation Flowchart}\label{fig:flowchart}
\end{figure}

The first version of the OxMat dataset was initiated and managed by Emeritus Professor Christopher Redman in the early 1990s. These data were used to develop the Oxford Dawes-Redman computerised CTG. Data acquired through this system have been reviewed weekly by clinical experts since its inception to ensure clinical accuracy. This system has been extensively reviewed elsewhere \cite{jones2022computerized}. The current version of OxMat is curated from six data sources, as shown in Figure \ref{fig:flowchart}. These include the Sonicaid Centrale CTG Viewing and Archiving System, the ORBIT data warehouse, the ViewPoint ultrasound system and the BadgerNet neonatal recording service within John Radcliffe Hospital and the MBRRACE national survey on neonatal deaths. The Sonicaid Centrale system contains CTGs acquired at the John Radcliffe Hospital using Huntleigh FM820E and FM830E, Philips HP50XM machines in the early period and Huntleigh Sonicaid Team 3s and Philips Avalon FA40s later on during the period covered in the dataset. All CTGs are acquired using the industry standard HP-50 communications protocol. This study was approved by the Ethics Committee in the Joint Research Office, Research and Development Department, Oxford University Hospitals NHS Trust: 13/SC/0153. The data collection protocol is also reported elsewhere \cite{jones2024performance}. \par

CTGs, patient information and perinatal data points were extracted from the previous two versions of the OxMat source. Specifically, OxMat-CTG provided the CTGs and anonymised maternal and neonatal clinical parameters, while OxMat-Patient provided the perinatal data points. OxMat-CTG comprises CTGs recorded in the Sonicaid Centrale system linked to anonymised neonatal, intrapartum, maternal and postpartum data collected from ORBIT. OxMat-Patient contains anonymised perinatal data points collected from ORBIT. Additional information about babies admitted to the Special Care Baby Unit (SCBU) was collected separately from ORBIT. The ultrasound data, additional neonatal data, and data on neonatal deaths were collected from ViewPoint, BadgerNet and MBRRACE, respectively. The raw dataset spans three decades, from 1991 to 2021, and contains a total of 779,017 data points imported from all sources. \par 

% \begin{figure}[h]
% \centering
% \includegraphics[width=.7\textwidth]{plot-nSingleton-nMultiple.png}
% \caption{Number of singleton vs multiple (2-6) deliveries}\label{fig:plot-nSingleton-nMultiple}
% \end{figure}

\begin{figure}[h]
\centering
\includegraphics[width=.8\textwidth]{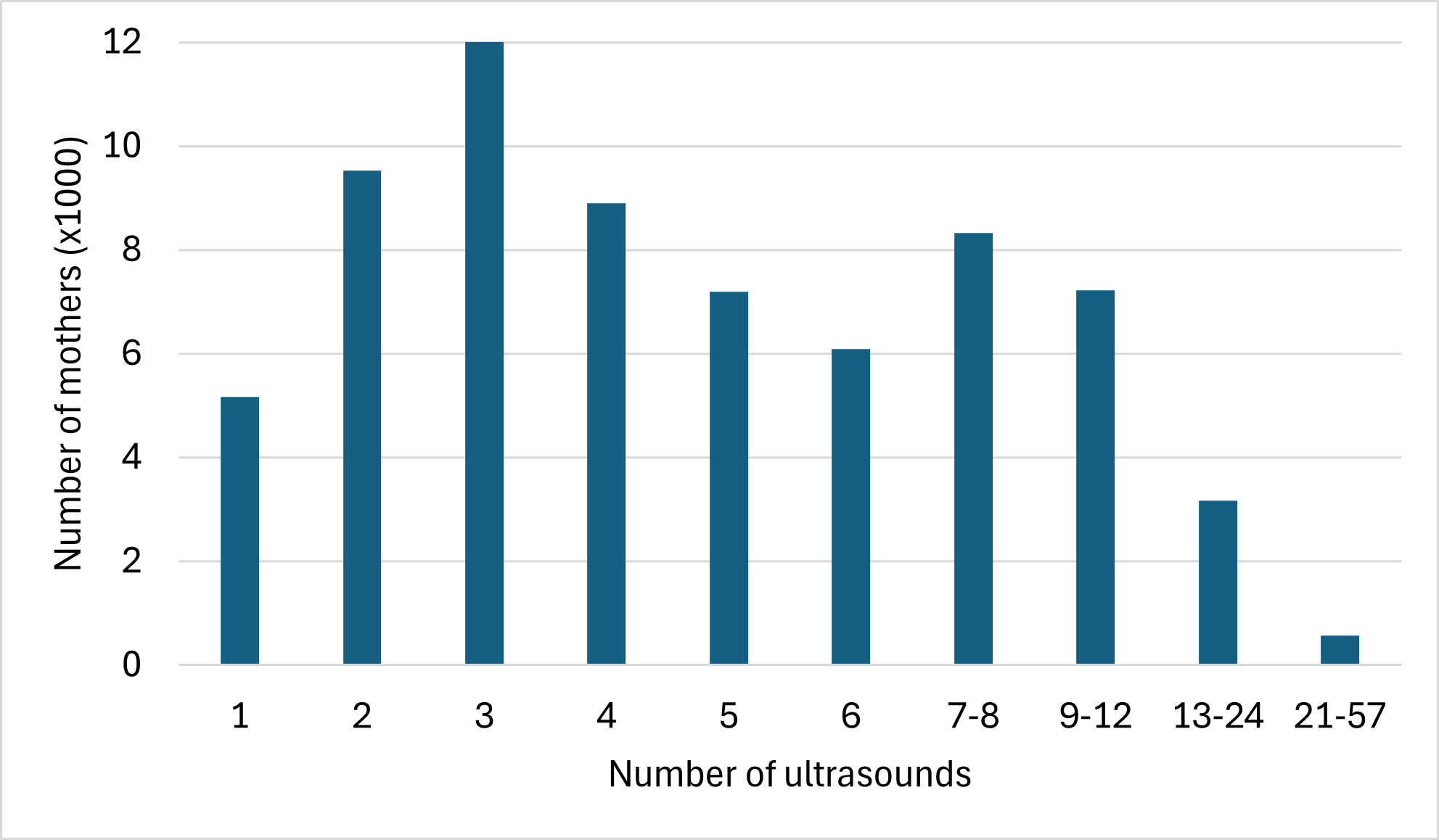}
\caption{Number of ultrasounds per mother}\label{fig:plot-nMoth-nUS}
\end{figure}

\section{Data Cleaning and Pre-Processing}\label{sec4}

Any identifiable information, such as patient names or NHS and hospital numbers, was removed from the raw dataset by clinical data managers working within the hospital. Anonymous numeric identifiers were allocated to each mother and baby. The data were then standardised to a uniform format. For example, by ensuring all entries for a categorical variable corresponded in a known way to a specific category. Entries that could not be standardised, for example an entry for a categorical variable that did not correspond to a known category, were excluded. Reports were generated to document missing values, necessitating an analysis of variable interdependencies and cases where missing values were expected (non-random missingness). For example, we expect labour onset information to be missing for a data point for mothers who did not labour. These were produced both for individual source datasets and for the merged data. Boolean columns were standardized to binary format, and any erroneous entries suggesting a missing value were discarded. A separate subset of the dataset was created focusing solely on singleton births, which account for 96.5\% pregnancies. This process culminated in the initial version of our dataset, comprising 810,017 data points for all births and 757,758 data points for only singleton births. \par

% record => change to data point => row in the dataset

Where applicable, data from all sources were merged using the baby's assigned anonymous study ID to prevent redundant data entries; if this was not available, the mother's was used. Subsequently, clinical variables were standardized, and each variable was examined for bespoke cleaning needs, such as outlier detection and adjustment, potential combination with similar variables, or recalculation from other data points. Transformations such as gestational age binning or ratio expressions for fetal measurements were also considered. Specific corrections for errors, such as re-encoding Boolean columns, were made. Imputation was performed for unexplained missing values, guided by inter-variable relationships to infer some missing values. The entire cleaning process was automated via a data dictionary, which outlined definitions, units, and prescribed cleaning for each variable. The data curation steps are illustrated in Figure \ref{fig:flowchart}. \par 

\begin{figure}[h]
\centering
\includegraphics[width=.6\textwidth]{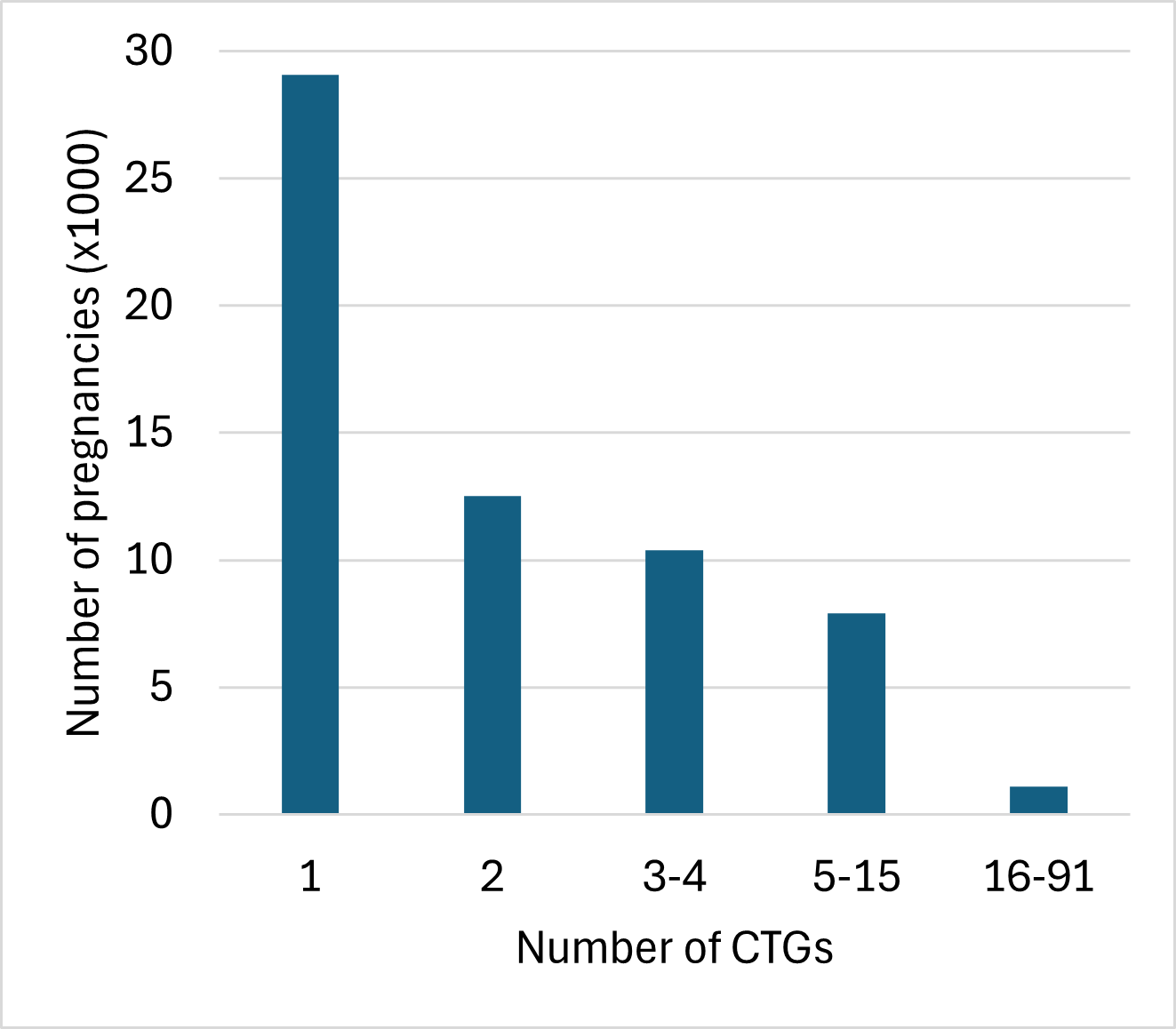}
\caption{Number of CTGs per pregnancy}\label{fig:plot-nPreg-nCTG}
\end{figure}

\begin{figure}[h]
\centering
\includegraphics[width=.6\textwidth]{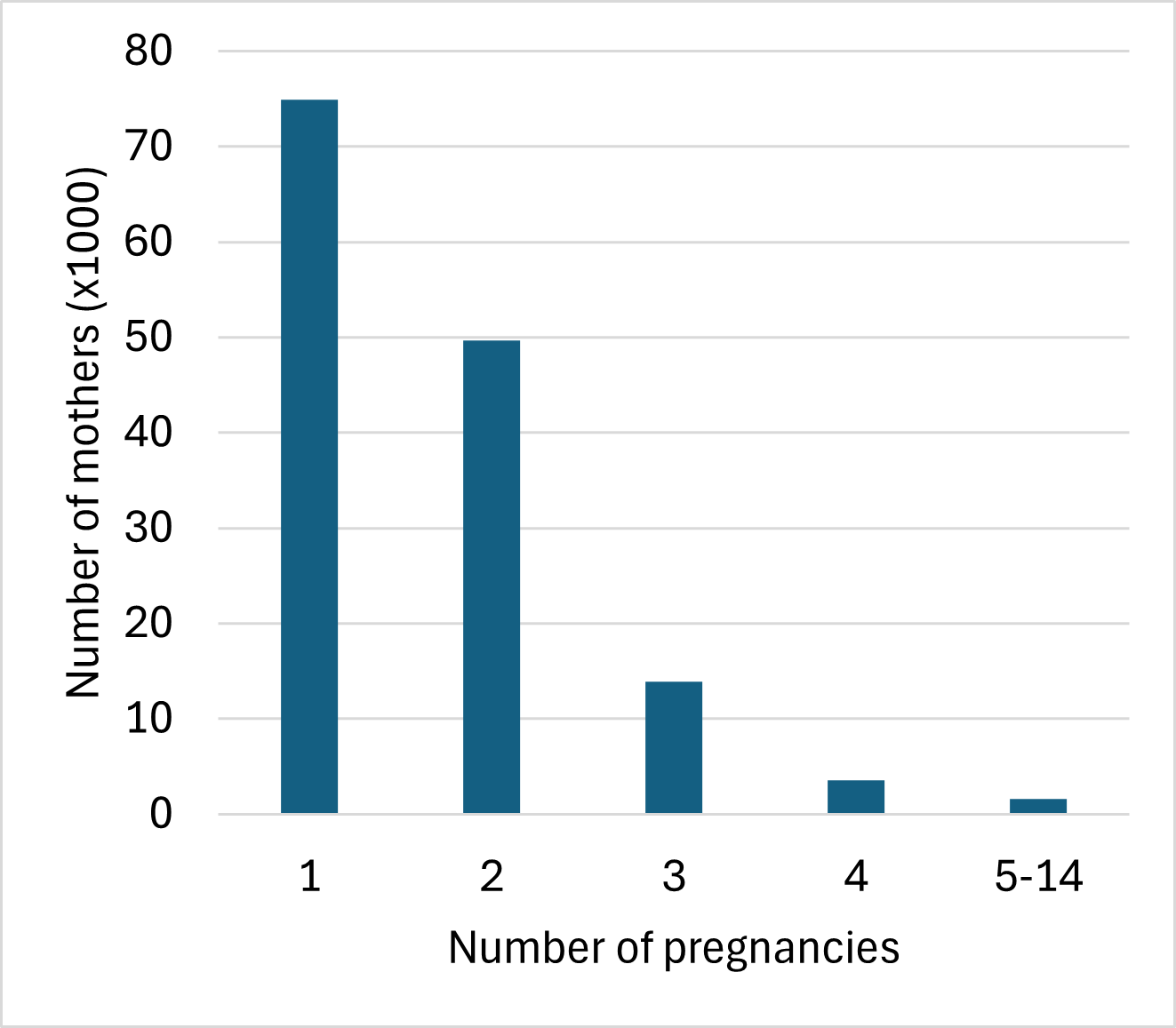}
\caption{Number of pregnancies per mother}\label{fig:plot-nMoth-nPreg}
\end{figure}

\begin{figure}[h]
\centering
\includegraphics[width=\textwidth]{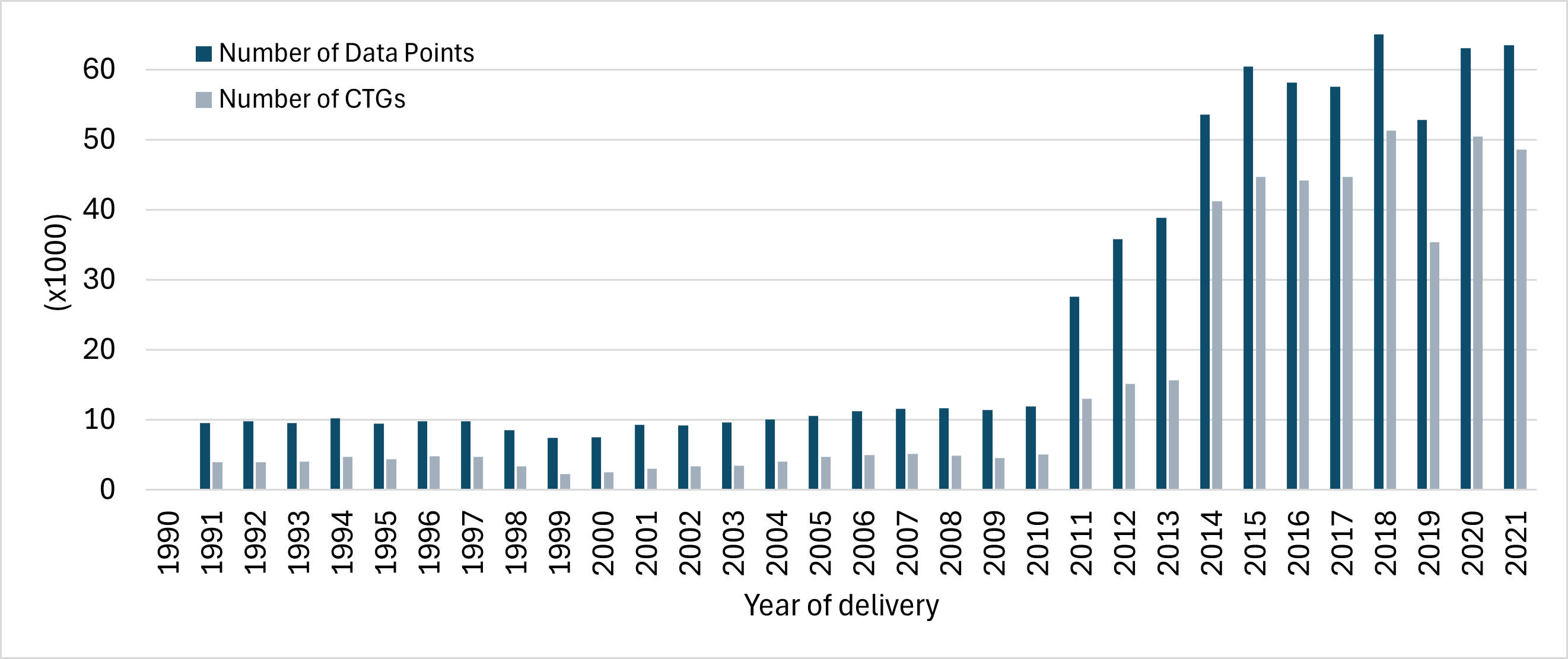}
\caption{Yearly breakdown of the number of total data points and CTGs in the final dataset}\label{fig:finalyearly-records-CTGs}
\end{figure}

% \begin{figure}[h]
% \centering
% \includegraphics[width=\textwidth]{plot-finalyearlystats2.png}
% \caption{Yearly breakdown of the number of mothers and babies in the final dataset}\label{fig:finalyearly-mother-baby}
% \end{figure}

\begin{figure}[h]
\centering
\includegraphics[width=\textwidth]{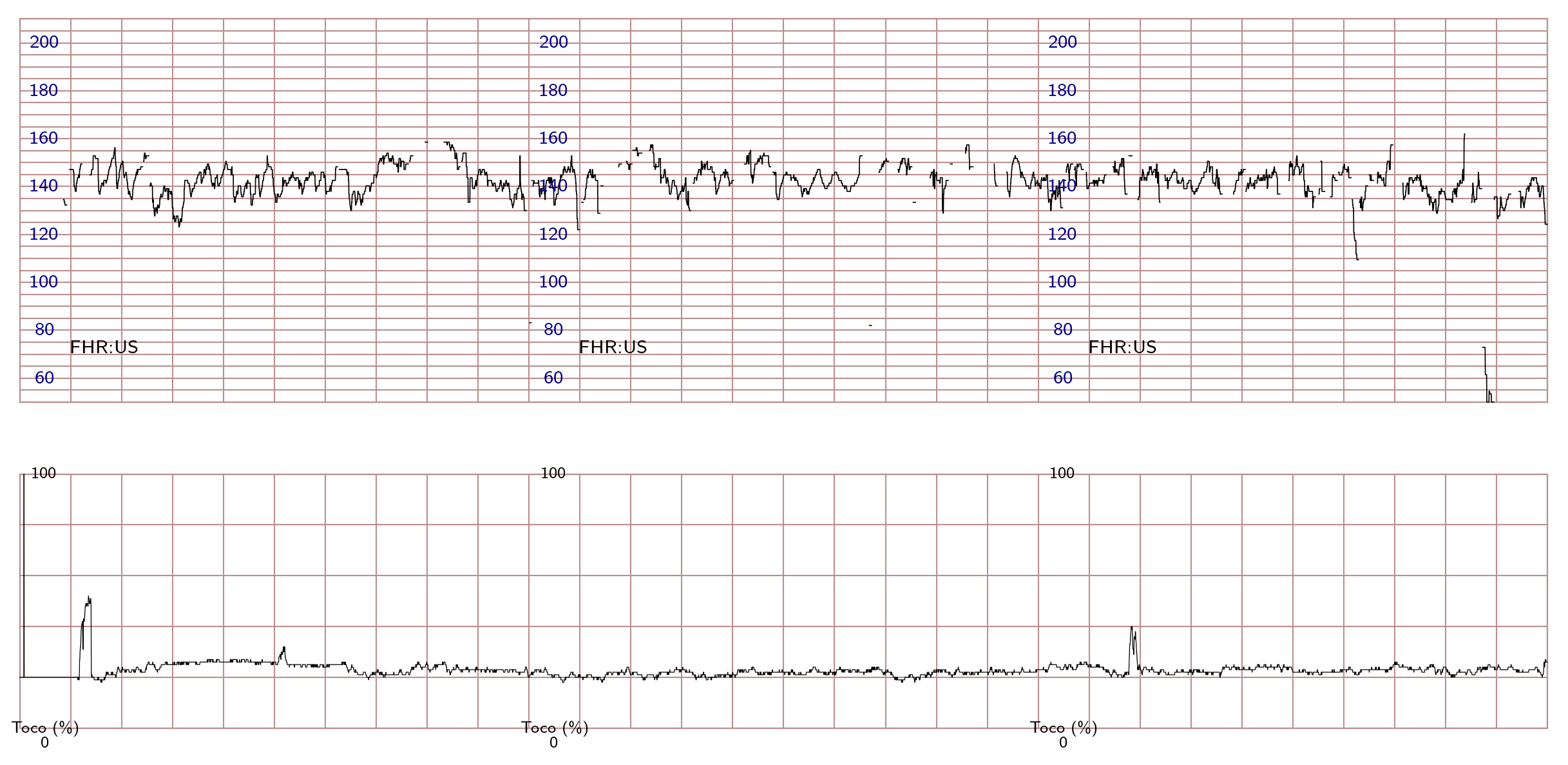}
\caption{A sample reassuring CTG trace with 3 accelerations and 2 decelerations, with a baseline FHR of 137 bpm and high and low short-term variation.}
\label{fig:ctg_reassuring}
\end{figure} 

% \begin{figure}[h]
% \centering
% \includegraphics[width=\textwidth]{new_reassuring_2.png}
% \caption{A sample reassuring CTG trace with 7 accelerations and no decelerations, with a baseline fetal heart rate 136 bpm and high and low short-term variation.}
% \label{fig:ctg_reassuring2}
% \end{figure} 

\begin{figure}[h]
\centering
\includegraphics[width=\textwidth]{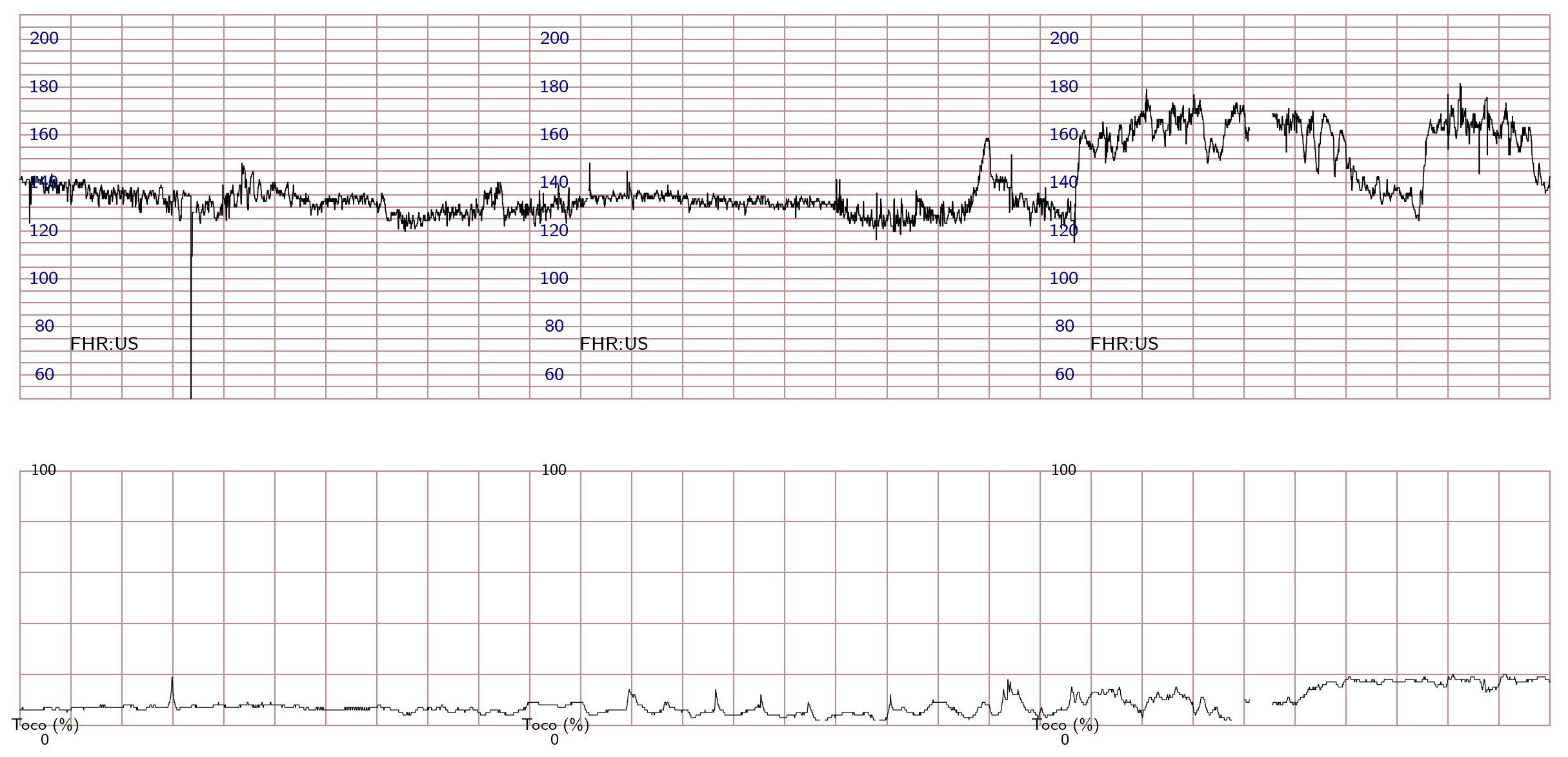}
\caption{A sample non-reassuring CTG trace with less than 8 out of 10 Dawes-Redman criteria (Table \ref{tab:dr_criteria}) met, taken 21 hours before an emergency Caesarean delivery before the onset of labour, resulting in low Apgar scores (one-minute Apgar score of 2) prompting successful resuscitation.}
\label{fig:ctg_nonreassuring}
\end{figure}
% A sample non-reassuring CTG trace with a Dawes-Redman score below 8/10, taken within 24h before an emergency caesarean delivery resulting in low Apgar scores for the newborn.

This process resulted in the final dataset comprising 1,689,503 data points for all births,  1,508,092 data points for only singleton births, and 295 variables. The final dataset contains 177,211 unique CTG recordings from 51,036 pregnancies. Additionally, three substantial auxiliary datasets were generated to document attendance, prescriptions, and clinical codes (International Statistical Classification of Diseases and Related Health Problems; ICD)  related to the patients. Due to their extensive size, these datasets remain separate, with the option to integrate them with the main dataset as future requirements may necessitate. A one-to-one correspondence was established between the unique identifiers and the individual mothers and babies. However, the relationship between these identifiers and the respective data points manifested as a one-to-many relationship. As illustrated in Figure \ref{fig:plot-nMoth-nUS} and Figure \ref{fig:plot-nPreg-nCTG} respectively, a significant proportion of women (92.4\%) underwent multiple ultrasound examinations, and more than half of pregnancies (52.3\%) were documented with several CTG recordings. Also, a considerable proportion of mothers (47.8\%) had multiple pregnancies, as shown in Figure \ref{fig:plot-nMoth-nPreg}. When two datasets, characterised by identifiers with a one-to-many relationship to the data point count, are merged, the resultant dataset invariably exhibits an increase in the total number of rows compared to the individual datasets. For instance, when a set of two data points of a baby's stay in the neonatal unit is merged with a set of three CTG recordings for the same pregnancy, the merged set will consist of six rows. \par 

The yearly breakdown of the total numbers of data points and CTGs in the final dataset is shown in Figure \ref{fig:finalyearly-records-CTGs}, which collectively illustrates the composition of the dataset over time. From 1991 to 2021, the dataset expanded annually by an average of 5,711 unique CTGs, originating from 1,646 pregnancies. This data predominantly consists of 96.46\% singleton births and 94.23\% antepartum CTGs. Figures \ref{fig:ctg_reassuring} and \ref{fig:ctg_nonreassuring} present examples of 30-minute CTG recordings. These figures show two key measurements: the FHR (shown in the upper part of each plot) and maternal tocometry (shown in the lower part). Figure \ref{fig:ctg_reassuring} shows a reassuring CTG trace, where the DR criteria, applied to the entire hour of recording, identified three accelerations and two decelerations. This trace also demonstrates an acceptable baseline FHR (137 beats per minute) and indicates both high and low short-term FHR variability. Conversely, Figure \ref{fig:ctg_nonreassuring} shows a non-reassuring CTG trace, characterized by a low DR score (below 8 out of 10). This trace was recorded 21 hours before an emergency Caesarean section before the onset of labour due to the concerning FHR pattern, and the newborn received low Apgar scores (1-min Apgar score of 2), indicating distress at birth. \par 

\section{Clinical Data}\label{sec5}

The OxMat dataset comprises a curated collection of clinical variables related to and contextualising the CTG data in the dataset. Its primary focus is to facilitate the analysis and interpretation of CTG data, thereby aiding in the early detection of fetal distress and improving pregnancy outcomes. The dataset is structured to include a wide range of clinical information about mothers and babies across multiple stages of pregnancy from multiple data sources. It is distinct in its comprehensive inclusion of a wide variety of variables, from administrative details to specific clinical information. Each variable is categorized based on its data source, description, stage of pregnancy, and the patient it pertains to, i.e. the mother or the fetus. The variables cover a broad spectrum of data points, including but not limited to demographic information, medical history, details of the pregnancy course, labour and delivery information, and postnatal outcomes for both the mother and the baby. Furthermore, the dataset is organized to allow for a granular analysis based on the stage of pregnancy, which includes antepartum, intrapartum, and postpartum phases, as well as based on the source patient, thereby offering a detailed view into the health status and medical interventions related to both the mother and the fetus. This structured approach enhances the usability of the dataset for specific research problems and also facilitates a comprehensive understanding of the factors influencing maternal and fetal outcomes. \par

The two OxMat sources that feed into this dataset (pre- and post-2011) are composed of 157 clinical variables spanning various stages of pregnancy, offering insights into a wide spectrum of clinical scenarios. The variables related to the neonate stage highlight a focused attention on postnatal outcomes and conditions: the baby's health immediately after birth, clinical interventions undertaken and follow-up outcomes until the end of SCBU stay. Moreover, the dataset comprises antepartum and intrapartum variables that are essential for monitoring fetal well-being, assessing the risk of complications and making informed clinical decisions during labour and delivery. The antepartum data, for example, includes variables related to maternal health assessments, fetal movements, and ultrasound findings, which are crucial for the early detection of potential issues. Variables from the intrapartum phase cover labour and delivery metrics, such as FHR patterns, maternal contractions and the presence of any complications during delivery. The inclusion of variables from both stages ensures a comprehensive timeline of maternal-fetal health that is critical for understanding the full spectrum of pregnancy care. The dataset's contribution extends to postpartum data, covering the newborn's health status, including Apgar scores, neonatal resuscitation measures, and admission to specialized care units. Such data is invaluable for evaluating the immediate impact of prenatal and intrapartum care strategies on neonatal outcomes. Additionally, OxMat provides maternal postpartum variables that assess the mother's health following delivery, ensuring that the dataset offers a holistic view of both maternal and neonatal well-being. Other variable types cover the whole pregnancy timeline, for example fields listing medications that were prescribed at every stage. \par 

Data originating from the BadgerNet source comprised 80 variables, primarily focused on the neonatal period. This collection is especially rich in details of the clinical care and outcomes of newborns, covering neonatal assessments, interventions and the subsequent monitoring that is critical in the immediate post-birth period. The variables provide exhaustive details on neonatal health, including important statistics such as birth weight and length, Apgar scores and the need for any resuscitation at birth. Furthermore, it includes comprehensive variables on neonatal intensive care admissions, detailing the duration of stay, treatments administered such as ventilation or Continuous Positive Airway Pressure (CPAP), and specific neonatal conditions diagnosed during the stay. The inclusion of BadgerNet data serves to bridge the gap between birth outcomes and neonatal care and to study the effectiveness of perinatal care and its impact on immediate neonatal health. It catalogues a detailed trajectory of the neonate's journey through critical care, encapsulating data on specialized interventions such as cooling therapy for hypoxic-ischemic encephalopathy, surgical procedures undertaken, and the outcomes of these interventions. The BadgerNet source contributes to a holistic view of maternal and neonatal health by including data on maternal health parameters post-delivery, maternal interventions during the neonatal period and the administrative aspects of neonatal care. This integrated approach allows for a comprehensive analysis of the continuum of care, from prenatal stages through to postnatal outcomes, highlighting the interdependencies between maternal health conditions, delivery practices, and neonatal health outcomes. This data is invaluable for researchers to examine neonatal treatments, the complexities of care in SCBU and the critical first steps in the lives of newborns. \par 

The inclusion of data from SCBU, ViewPoint and MBRRACE sources in this dataset introduces a multifaceted perspective on maternal and neonatal health, enriching it with a diverse array of clinical variables that span the entire spectrum of perinatal care. The SCBU data, though concise with 3 variables, is critical as it focuses on the neonatal intensive care aspect, providing insights into the prevalence and duration of specialized care required by newborns. This segment of data is pivotal for understanding the complexities and demands of neonatal care in the special care units. ViewPoint contributed 46 variables comprising detailed ultrasound findings, maternal health evaluations during and after pregnancy, and critical parameters that influence pregnancy outcomes. ViewPoint provides a longitudinal view of maternal-fetal health, enabling researchers to trace the progression of pregnancy, identify potential complications early, and understand the impact of prenatal care on both maternal and fetal outcomes. MBRRACE, with 12 variables centred predominantly on adverse neonatal outcomes, covers the mortality and morbidity of newborns. It includes detailed codification of causes of death, enhancing the dataset’s utility for epidemiological studies focused on neonatal mortality and morbidity trends. This data is crucial for identifying risk factors associated with adverse neonatal outcomes, enabling healthcare providers to implement targeted interventions aimed at mitigating these risks. \par

In summary, the OxMat dataset presents a groundbreaking compilation of clinical variables sourced from OxMat-CTG, OxMat-Patient, BadgerNet, ViewPoint, and MBRRACE, offering an unparalleled depth and breadth of data for the maternal-fetal health domain. Organizing these variables across different stages of pregnancy and patient sources enables a fine-grained analysis of maternal and fetal health, from prenatal to postnatal phases. The diverse origins of the data enrich the dataset, providing a comprehensive view that encompasses antepartum assessments, intrapartum management, neonatal care and postpartum outcomes. This rich dataset facilitates advanced research in clinical AI and paves the way for data-driven approaches to enhance our understanding of maternal and fetal health, aiming for improved outcomes and the well-being of mothers and babies. 

\section{Conclusion}\label{sec6}
The paper presents a comprehensive review of the existing CTG datasets in fetal monitoring research, highlighting the limitations in volume, clinical detail, and stage-specific focus. The new Oxford Maternity (OxMat) dataset presented in this paper stands as the most extensive curated collection of CTG recordings, uniquely featuring raw time series data for machine learning, coupled with in-depth clinical data for both mothers and infants. Addressing a significant gap in women's health data science and electronic fetal monitoring, the OxMat dataset comprises 177,211 unique CTG recordings from 51,036 pregnancies and more than 200 clinical variables per data point. The careful curation and comprehensive review by clinical experts ensure near-complete data accuracy for essential outcomes, such as stillbirths or neonatal fatalities. This contribution paves the way for novel AI techniques to improve maternal and fetal outcomes through advanced CTG analysis and interpretation.

\section*{Acknowledgements}
This work was supported by the Medical Research Council [MR/X029689/1] and The Alan Turing Institute's Enrichment Scheme.

This work is indebted to the work of Profesor Christopher Redman on developing the Dawes-Redman criteria and advancing CTGs at the University of Oxford; and to Pawe\l{} Szafranski and James Bland, data managers within the University of Oxford, for their work compiling and anonymising the original data.

\bibliography{references}% common bib file


\begin{thebibliography}{36}
\ifx \bisbn   \undefined \def \bisbn  #1{ISBN #1}\fi
\ifx \binits  \undefined \def \binits#1{#1}\fi
\ifx \bauthor  \undefined \def \bauthor#1{#1}\fi
\ifx \batitle  \undefined \def \batitle#1{#1}\fi
\ifx \bjtitle  \undefined \def \bjtitle#1{#1}\fi
\ifx \bvolume  \undefined \def \bvolume#1{\textbf{#1}}\fi
\ifx \byear  \undefined \def \byear#1{#1}\fi
\ifx \bissue  \undefined \def \bissue#1{#1}\fi
\ifx \bfpage  \undefined \def \bfpage#1{#1}\fi
\ifx \blpage  \undefined \def \blpage #1{#1}\fi
\ifx \burl  \undefined \def \burl#1{\textsf{#1}}\fi
\ifx \doiurl  \undefined \def \doiurl#1{\url{https://doi.org/#1}}\fi
\ifx \betal  \undefined \def \betal{\textit{et al.}}\fi
\ifx \binstitute  \undefined \def \binstitute#1{#1}\fi
\ifx \binstitutionaled  \undefined \def \binstitutionaled#1{#1}\fi
\ifx \bctitle  \undefined \def \bctitle#1{#1}\fi
\ifx \beditor  \undefined \def \beditor#1{#1}\fi
\ifx \bpublisher  \undefined \def \bpublisher#1{#1}\fi
\ifx \bbtitle  \undefined \def \bbtitle#1{#1}\fi
\ifx \bedition  \undefined \def \bedition#1{#1}\fi
\ifx \bseriesno  \undefined \def \bseriesno#1{#1}\fi
\ifx \blocation  \undefined \def \blocation#1{#1}\fi
\ifx \bsertitle  \undefined \def \bsertitle#1{#1}\fi
\ifx \bsnm \undefined \def \bsnm#1{#1}\fi
\ifx \bsuffix \undefined \def \bsuffix#1{#1}\fi
\ifx \bparticle \undefined \def \bparticle#1{#1}\fi
\ifx \barticle \undefined \def \barticle#1{#1}\fi
\bibcommenthead
\ifx \bconfdate \undefined \def \bconfdate #1{#1}\fi
\ifx \botherref \undefined \def \botherref #1{#1}\fi
\ifx \url \undefined \def \url#1{\textsf{#1}}\fi
\ifx \bchapter \undefined \def \bchapter#1{#1}\fi
\ifx \bbook \undefined \def \bbook#1{#1}\fi
\ifx \bcomment \undefined \def \bcomment#1{#1}\fi
\ifx \oauthor \undefined \def \oauthor#1{#1}\fi
\ifx \citeauthoryear \undefined \def \citeauthoryear#1{#1}\fi
\ifx \endbibitem  \undefined \def \endbibitem {}\fi
\ifx \bconflocation  \undefined \def \bconflocation#1{#1}\fi
\ifx \arxivurl  \undefined \def \arxivurl#1{\textsf{#1}}\fi
\csname PreBibitemsHook\endcsname

%%% 1
\bibitem[\protect\citeauthoryear{}{}]{market2023}
\begin{botherref}
{AI In Healthcare Market Size, Share {\&} Growth Report, 2030}.
\url{https://www.grandviewresearch.com/industry-analysis/artificial-intelligence-ai-healthcare-market}
\end{botherref}
\endbibitem

%%% 2
\bibitem[\protect\citeauthoryear{}{}]{amazon2023}
\begin{botherref}
{Amazon's {\$}4B investment moves it deeper into healthcare AI}.
\url{https://www.fiercehealthcare.com/health-tech/how-amazons-4b-investment-ai-company-anthropic-impacts-healthcare}
\end{botherref}
\endbibitem

%%% 3
\bibitem[\protect\citeauthoryear{}{}]{corti2023}
\begin{botherref}
{Corti, an AI ‘co-pilot’ for healthcare clinicians, raises {\$}60M |
  TechCrunch}.
\url{https://techcrunch.com/2023/09/20/corti-an-ai-co-pilot-for-healthcare-clinicians-raises-60m/?guccounter=1}
\end{botherref}
\endbibitem

%%% 4
\bibitem[\protect\citeauthoryear{}{}]{philips2023}
\begin{botherref}
{Philips develops AI ultrasound to expand maternal health - News | Philips}.
\url{https://www.philips.com/a-w/about/news/archive/standard/news/press/2023/20231107-philips-program-developing-ai-powered-ultrasound-to-expand-access-to-maternal-health-receives-major-funding-boost.html}
\end{botherref}
\endbibitem

%%% 5
\bibitem[\protect\citeauthoryear{Alowais
  et~al.}{2023}]{alowais2023revolutionizing}
\begin{barticle}
\bauthor{\bsnm{Alowais}, \binits{S.A.}},
\bauthor{\bsnm{Alghamdi}, \binits{S.S.}},
\bauthor{\bsnm{Alsuhebany}, \binits{N.}},
\bauthor{\bsnm{Alqahtani}, \binits{T.}},
\bauthor{\bsnm{Alshaya}, \binits{A.I.}},
\bauthor{\bsnm{Almohareb}, \binits{S.N.}},
\bauthor{\bsnm{Aldairem}, \binits{A.}},
\bauthor{\bsnm{Alrashed}, \binits{M.}},
\bauthor{\bsnm{Bin~Saleh}, \binits{K.}},
\bauthor{\bsnm{Badreldin}, \binits{H.A.}},
\bauthor{\bsnm{Al~Yami}, \binits{M.S.}},
\bauthor{\bsnm{Al~Harbi}, \binits{S.}},
\bauthor{\bsnm{Albekairy}, \binits{A.M.}}:
\batitle{{Revolutionizing healthcare: the role of artificial intelligence in
  clinical practice}}.
\bjtitle{BMC Medical Education 2023 23:1}
\bvolume{23}(\bissue{1}),
\bfpage{1}--\blpage{15}
(\byear{2023})
\doiurl{10.1186/S12909-023-04698-Z}
\end{barticle}
\endbibitem

%%% 6
\bibitem[\protect\citeauthoryear{Sonawani
  et~al.}{2023}]{sonawani2023biomedical}
\begin{barticle}
\bauthor{\bsnm{Sonawani}, \binits{S.}},
\bauthor{\bsnm{Patil}, \binits{K.}},
\bauthor{\bsnm{Natarajan}, \binits{P.}}:
\batitle{{Biomedical signal processing for health monitoring applications: a
  review}}.
\bjtitle{International Journal of Applied Systemic Studies}
\bvolume{10}(\bissue{1}),
\bfpage{44}
(\byear{2023})
\doiurl{10.1504/IJASS.2023.129065}
\end{barticle}
\endbibitem

%%% 7
\bibitem[\protect\citeauthoryear{Qureshi et~al.}{2023}]{qureshi2023artificial}
\begin{barticle}
\bauthor{\bsnm{Qureshi}, \binits{R.}},
\bauthor{\bsnm{Irfan}, \binits{M.}},
\bauthor{\bsnm{Ali}, \binits{H.}},
\bauthor{\bsnm{Khan}, \binits{A.}},
\bauthor{\bsnm{Nittala}, \binits{A.S.}},
\bauthor{\bsnm{Ali}, \binits{S.}},
\bauthor{\bsnm{Shah}, \binits{A.}},
\bauthor{\bsnm{Gondal}, \binits{T.M.}},
\bauthor{\bsnm{Sadak}, \binits{F.}},
\bauthor{\bsnm{Shah}, \binits{Z.}},
\bauthor{\bsnm{Hadi}, \binits{M.U.}},
\bauthor{\bsnm{Khan}, \binits{S.}},
\bauthor{\bsnm{Al-Tashi}, \binits{Q.}},
\bauthor{\bsnm{Wu}, \binits{J.}},
\bauthor{\bsnm{Bermak}, \binits{A.}},
\bauthor{\bsnm{Alam}, \binits{T.}}:
\batitle{{Artificial Intelligence and Biosensors in Healthcare and Its Clinical
  Relevance: A Review}}.
\bjtitle{IEEE Access}
\bvolume{11},
\bfpage{61600}--\blpage{61620}
(\byear{2023})
\doiurl{10.1109/ACCESS.2023.3285596}
\end{barticle}
\endbibitem

%%% 8
\bibitem[\protect\citeauthoryear{Manickam
  et~al.}{2022}]{manickam2022artificial}
\begin{botherref}
\oauthor{\bsnm{Manickam}, \binits{P.}},
\oauthor{\bsnm{Mariappan}, \binits{S.A.}},
\oauthor{\bsnm{Murugesan}, \binits{S.M.}},
\oauthor{\bsnm{Hansda}, \binits{S.}},
\oauthor{\bsnm{Kaushik}, \binits{A.}},
\oauthor{\bsnm{Shinde}, \binits{R.}},
\oauthor{\bsnm{Thipperudraswamy}, \binits{S.P.}}:
{Artificial Intelligence (AI) and Internet of Medical Things (IoMT) Assisted
  Biomedical Systems for Intelligent Healthcare}.
Biosensors
\textbf{12}(8)
(2022)
\doiurl{10.3390/BIOS12080562}
\end{botherref}
\endbibitem

%%% 9
\bibitem[\protect\citeauthoryear{Junaid et~al.}{2022}]{junaid2022recent}
\begin{barticle}
\bauthor{\bsnm{Junaid}, \binits{S.B.}},
\bauthor{\bsnm{Imam}, \binits{A.A.}},
\bauthor{\bsnm{Abdulkarim}, \binits{M.}},
\bauthor{\bsnm{Surakat}, \binits{Y.A.}},
\bauthor{\bsnm{Balogun}, \binits{A.O.}},
\bauthor{\bsnm{Kumar}, \binits{G.}},
\bauthor{\bsnm{Shuaibu}, \binits{A.N.}},
\bauthor{\bsnm{Garba}, \binits{A.}},
\bauthor{\bsnm{Sahalu}, \binits{Y.}},
\bauthor{\bsnm{Mohammed}, \binits{A.}},
\bauthor{\bsnm{Mohammed}, \binits{T.Y.}},
\bauthor{\bsnm{Abdulkadir}, \binits{B.A.}},
\bauthor{\bsnm{Abba}, \binits{A.A.}},
\bauthor{\bsnm{Iliyasu~Kakumi}, \binits{N.A.}},
\bauthor{\bsnm{Hashim}, \binits{A.S.}}:
\batitle{{Recent Advances in Artificial Intelligence and Wearable Sensors in
  Healthcare Delivery}}.
\bjtitle{Applied Sciences 2022, Vol. 12, Page 10271}
\bvolume{12}(\bissue{20}),
\bfpage{10271}
(\byear{2022})
\doiurl{10.3390/APP122010271}
\end{barticle}
\endbibitem

%%% 10
\bibitem[\protect\citeauthoryear{Choy et~al.}{2018}]{choy2018current}
\begin{barticle}
\bauthor{\bsnm{Choy}, \binits{G.}},
\bauthor{\bsnm{Khalilzadeh}, \binits{O.}},
\bauthor{\bsnm{Michalski}, \binits{M.}},
\bauthor{\bsnm{Do}, \binits{S.}},
\bauthor{\bsnm{Samir}, \binits{A.E.}},
\bauthor{\bsnm{Pianykh}, \binits{O.S.}},
\bauthor{\bsnm{Geis}, \binits{J.R.}},
\bauthor{\bsnm{Pandharipande}, \binits{P.V.}},
\bauthor{\bsnm{Brink}, \binits{J.A.}},
\bauthor{\bsnm{Dreyer}, \binits{K.J.}}:
\batitle{{Current applications and future impact of machine learning in
  radiology}}.
\bjtitle{Radiology}
\bvolume{288}(\bissue{2}),
\bfpage{318}--\blpage{328}
(\byear{2018})
\doiurl{10.1148/RADIOL.2018171820/ASSET/IMAGES/LARGE/RADIOL.2018171820.FIG8.JPEG}
\end{barticle}
\endbibitem

%%% 11
\bibitem[\protect\citeauthoryear{Rodriguez-Ruiz
  et~al.}{2019}]{rodriguez2019stand}
\begin{barticle}
\bauthor{\bsnm{Rodriguez-Ruiz}, \binits{A.}},
\bauthor{\bsnm{L{\aa}ng}, \binits{K.}},
\bauthor{\bsnm{Gubern-Merida}, \binits{A.}},
\bauthor{\bsnm{Broeders}, \binits{M.}},
\bauthor{\bsnm{Gennaro}, \binits{G.}},
\bauthor{\bsnm{Clauser}, \binits{P.}},
\bauthor{\bsnm{Helbich}, \binits{T.H.}},
\bauthor{\bsnm{Chevalier}, \binits{M.}},
\bauthor{\bsnm{Tan}, \binits{T.}},
\bauthor{\bsnm{Mertelmeier}, \binits{T.}},
\bauthor{\bsnm{Wallis}, \binits{M.G.}},
\bauthor{\bsnm{Andersson}, \binits{I.}},
\bauthor{\bsnm{Zackrisson}, \binits{S.}},
\bauthor{\bsnm{Mann}, \binits{R.M.}},
\bauthor{\bsnm{Sechopoulos}, \binits{I.}}:
\batitle{{Stand-Alone Artificial Intelligence for Breast Cancer Detection in
  Mammography: Comparison With 101 Radiologists}}.
\bjtitle{JNCI: Journal of the National Cancer Institute}
\bvolume{111}(\bissue{9}),
\bfpage{916}--\blpage{922}
(\byear{2019})
\doiurl{10.1093/JNCI/DJY222}
\end{barticle}
\endbibitem

%%% 12
\bibitem[\protect\citeauthoryear{Stirnemann
  et~al.}{2023}]{stirnemann2023development}
\begin{barticle}
\bauthor{\bsnm{Stirnemann}, \binits{J.J.}},
\bauthor{\bsnm{Besson}, \binits{R.}},
\bauthor{\bsnm{Spaggiari}, \binits{E.}},
\bauthor{\bsnm{Rojo}, \binits{S.}},
\bauthor{\bsnm{Loge}, \binits{F.}},
\bauthor{\bsnm{Peyro-Saint-Paul}, \binits{H.}},
\bauthor{\bsnm{Allassonniere}, \binits{S.}},
\bauthor{\bsnm{Le~Pennec}, \binits{E.}},
\bauthor{\bsnm{Hutchinson}, \binits{C.}},
\bauthor{\bsnm{Sebire}, \binits{N.}},
\bauthor{\bsnm{Ville}, \binits{Y.}}:
\batitle{{Development and clinical validation of real-time artificial
  intelligence diagnostic companion for fetal ultrasound examination}}.
\bjtitle{Ultrasound in Obstetrics {\&} Gynecology}
\bvolume{62}(\bissue{3}),
\bfpage{353}--\blpage{360}
(\byear{2023})
\doiurl{10.1002/UOG.26242}
\end{barticle}
\endbibitem

%%% 13
\bibitem[\protect\citeauthoryear{Morid et~al.}{2023}]{morid2023time}
\begin{botherref}
\oauthor{\bsnm{Morid}, \binits{M.A.}},
\oauthor{\bsnm{Sheng}, \binits{O.R.L.}},
\oauthor{\bsnm{Dunbar}, \binits{J.}}:
{Time Series Prediction Using Deep Learning Methods in Healthcare}.
ACM Transactions on Management Information Systems
\textbf{14}(1)
(2023)
\doiurl{10.1145/3531326}
\end{botherref}
\endbibitem

%%% 14
\bibitem[\protect\citeauthoryear{}{}]{wec2024}
\begin{botherref}
{Closing the Women’s Health Gap to Improve Lives and Economies | World
  Economic Forum}.
\url{https://www.weforum.org/publications/closing-the-women-s-health-gap-a-1-trillion-opportunity-to-improve-lives-and-economies/}
\end{botherref}
\endbibitem

%%% 15
\bibitem[\protect\citeauthoryear{Pardey et~al.}{2002}]{pardey2002computer}
\begin{barticle}
\bauthor{\bsnm{Pardey}, \binits{J.}},
\bauthor{\bsnm{Moulden}, \binits{M.}},
\bauthor{\bsnm{Redman}, \binits{C.W.G.}}:
\batitle{{A computer system for the numerical analysis of nonstress tests}}.
\bjtitle{American Journal of Obstetrics and Gynecology}
\bvolume{186}(\bissue{5}),
\bfpage{1095}--\blpage{1103}
(\byear{2002})
\doiurl{10.1067/MOB.2002.122447}
\end{barticle}
\endbibitem

%%% 16
\bibitem[\protect\citeauthoryear{Ayres-De-Campos et~al.}{2015}]{ayres2015figo}
\begin{barticle}
\bauthor{\bsnm{Ayres-De-Campos}, \binits{D.}},
\bauthor{\bsnm{Spong}, \binits{C.Y.}},
\bauthor{\bsnm{Chandraharan}, \binits{E.}}:
\batitle{{FIGO consensus guidelines on intrapartum fetal monitoring:
  Cardiotocography}}.
\bjtitle{International Journal of Gynecology {\&} Obstetrics}
\bvolume{131}(\bissue{1}),
\bfpage{13}--\blpage{24}
(\byear{2015})
\doiurl{10.1016/J.IJGO.2015.06.020}
\end{barticle}
\endbibitem

%%% 17
\bibitem[\protect\citeauthoryear{Alfirevic
  et~al.}{2017}]{alfirevic2017continuous}
\begin{botherref}
\oauthor{\bsnm{Alfirevic}, \binits{Z.}},
\oauthor{\bsnm{Devane}, \binits{D.}},
\oauthor{\bsnm{Gyte}, \binits{G.M.L.}},
\oauthor{\bsnm{Cuthbert}, \binits{A.}}:
{Continuous cardiotocography (CTG) as a form of electronic fetal monitoring
  (EFM) for fetal assessment during labour}.
Cochrane Database of Systematic Reviews
\textbf{2017}(2)
(2017)
\doiurl{10.1002/14651858.CD006066.PUB3/MEDIA/CDSR/CD006066/IMAGE{\_}N/NCD006066-CMP-002-02.}
\end{botherref}
\endbibitem

%%% 18
\bibitem[\protect\citeauthoryear{Al-yousif
  et~al.}{2021}]{alyousif2021systematic}
\begin{barticle}
\bauthor{\bsnm{Al-yousif}, \binits{S.}},
\bauthor{\bsnm{Jaenul}, \binits{A.}},
\bauthor{\bsnm{Al-Dayyeni}, \binits{W.}},
\bauthor{\bsnm{Alamoodi}, \binits{A.}},
\bauthor{\bsnm{Jabori}, \binits{I.}},
\bauthor{\bsnm{Tahir}, \binits{N.M.}},
\bauthor{\bsnm{Alrawi}, \binits{A.A.A.}},
\bauthor{\bsnm{C{\"{o}}mert}, \binits{Z.}},
\bauthor{\bsnm{Al-shareefi}, \binits{N.A.}},
\bauthor{\bsnm{Saleh}, \binits{A.H.}}:
\batitle{{A systematic review of automated pre-processing, feature extraction
  and classification of cardiotocography}}.
\bjtitle{PeerJ Computer Science}
\bvolume{7},
\bfpage{1}--\blpage{37}
(\byear{2021})
\doiurl{10.7717/PEERJ-CS.452}
\end{barticle}
\endbibitem

%%% 19
\bibitem[\protect\citeauthoryear{Aeberhard
  et~al.}{2024}]{aeberhard2024introducing}
\begin{barticle}
\bauthor{\bsnm{Aeberhard}, \binits{J.L.}},
\bauthor{\bsnm{Radan}, \binits{A.-P.}},
\bauthor{\bsnm{Soltani}, \binits{R.A.}},
\bauthor{\bsnm{Strahm}, \binits{K.M.}},
\bauthor{\bsnm{Schneider}, \binits{S.}},
\bauthor{\bsnm{Carri{\'{e}}}, \binits{A.}},
\bauthor{\bsnm{Lemay}, \binits{M.}},
\bauthor{\bsnm{Krauss}, \binits{J.}},
\bauthor{\bsnm{Delgado-Gonzalo}, \binits{R.}},
\bauthor{\bsnm{Surbek}, \binits{D.}}:
\batitle{{Introducing Artificial Intelligence in Interpretation of Foetal
  Cardiotocography: Medical Dataset Curation and Preliminary Coding—An
  Interdisciplinary Project}}.
\bjtitle{Methods and Protocols 2024, Vol. 7, Page 5}
\bvolume{7}(\bissue{1}),
\bfpage{5}
(\byear{2024})
\doiurl{10.3390/MPS7010005}
\end{barticle}
\endbibitem

%%% 20
\bibitem[\protect\citeauthoryear{Aeberhard
  et~al.}{2023}]{aeberhard2023artificial}
\begin{barticle}
\bauthor{\bsnm{Aeberhard}, \binits{J.L.}},
\bauthor{\bsnm{Radan}, \binits{A.P.}},
\bauthor{\bsnm{Delgado-Gonzalo}, \binits{R.}},
\bauthor{\bsnm{Strahm}, \binits{K.M.}},
\bauthor{\bsnm{Sigurthorsdottir}, \binits{H.B.}},
\bauthor{\bsnm{Schneider}, \binits{S.}},
\bauthor{\bsnm{Surbek}, \binits{D.}}:
\batitle{{Artificial intelligence and machine learning in cardiotocography: A
  scoping review}}.
\bjtitle{European Journal of Obstetrics {\&} Gynecology and Reproductive
  Biology}
\bvolume{281},
\bfpage{54}--\blpage{62}
(\byear{2023})
\doiurl{10.1016/J.EJOGRB.2022.12.008}
\end{barticle}
\endbibitem

%%% 21
\bibitem[\protect\citeauthoryear{Barnova et~al.}{2024}]{barnova2024artificial}
\begin{botherref}
\oauthor{\bsnm{Barnova}, \binits{K.}},
\oauthor{\bsnm{Martinek}, \binits{R.}},
\oauthor{\bsnm{Vilimkova~Kahankova}, \binits{R.}},
\oauthor{\bsnm{Jaros}, \binits{R.}},
\oauthor{\bsnm{Snasel}, \binits{V.}},
\oauthor{\bsnm{Mirjalili}, \binits{S.}}:
{Artificial Intelligence and Machine Learning in Electronic Fetal Monitoring}.
Archives of Computational Methods in Engineering 2024,
1--32
(2024)
\doiurl{10.1007/S11831-023-10055-6}
\end{botherref}
\endbibitem

%%% 22
\bibitem[\protect\citeauthoryear{Brocklehurst
  et~al.}{2017}]{brocklehurst2017computerised}
\begin{barticle}
\bauthor{\bsnm{Brocklehurst}, \binits{P.}},
\bauthor{\bsnm{Field}, \binits{D.}},
\bauthor{\bsnm{Greene}, \binits{K.}},
\bauthor{\bsnm{Juszczak}, \binits{E.}},
\bauthor{\bsnm{Kenyon}, \binits{S.}},
\bauthor{\bsnm{Linsell}, \binits{L.}},
\bauthor{\bsnm{Mabey}, \binits{C.}},
\bauthor{\bsnm{Newburn}, \binits{M.}},
\bauthor{\bsnm{Plachcinski}, \binits{R.}},
\bauthor{\bsnm{Quigley}, \binits{M.}},
\bauthor{\bsnm{Schroeder}, \binits{E.}},
\bauthor{\bsnm{Steer}, \binits{P.}},
\bauthor{\bsnm{Keith}, \binits{R.}},
\bauthor{\bsnm{Johns}, \binits{N.}},
\bauthor{\bsnm{Johnston}, \binits{T.}},
\bauthor{\bsnm{Barnfield}, \binits{G.}},
\bauthor{\bsnm{Davies}, \binits{K.}},
\bauthor{\bsnm{Johnson}, \binits{M.}},
\bauthor{\bsnm{Patterson}, \binits{H.}},
\bauthor{\bsnm{Montague}, \binits{I.}},
\bauthor{\bsnm{Watmore}, \binits{S.}},
\bauthor{\bsnm{Stolton}, \binits{A.}},
\bauthor{\bsnm{Parisaei}, \binits{M.}},
\bauthor{\bsnm{McGhee}, \binits{N.}},
\bauthor{\bsnm{Segovia}, \binits{S.}},
\bauthor{\bsnm{Martindale}, \binits{E.}},
\bauthor{\bsnm{Jackson}, \binits{H.}},
\bauthor{\bsnm{Holleran}, \binits{J.}},
\bauthor{\bsnm{Roberts}, \binits{D.}},
\bauthor{\bsnm{Holt}, \binits{S.}},
\bauthor{\bsnm{Dragovic}, \binits{B.}},
\bauthor{\bsnm{Willmott-Powell}, \binits{M.}},
\bauthor{\bsnm{Hutchinson}, \binits{L.}},
\bauthor{\bsnm{Toth}, \binits{B.}},
\bauthor{\bsnm{Chandler}, \binits{G.}},
\bauthor{\bsnm{Ridley}, \binits{S.}},
\bauthor{\bsnm{Bugg}, \binits{G.}},
\bauthor{\bsnm{Molnar}, \binits{A.}},
\bauthor{\bsnm{Lochrie}, \binits{D.}},
\bauthor{\bsnm{Connor}, \binits{J.}},
\bauthor{\bsnm{Howe}, \binits{D.}},
\bauthor{\bsnm{Head}, \binits{K.}},
\bauthor{\bsnm{Wellstead}, \binits{S.}},
\bauthor{\bsnm{Mathers}, \binits{A.}},
\bauthor{\bsnm{Walker}, \binits{L.}},
\bauthor{\bsnm{Crawford}, \binits{I.}},
\bauthor{\bsnm{Davies}, \binits{D.}},
\bauthor{\bsnm{Garner}, \binits{Z.}},
\bauthor{\bsnm{Galloway}, \binits{L.}},
\bauthor{\bsnm{Davies}, \binits{Y.}},
\bauthor{\bsnm{Smith}, \binits{C.}},
\bauthor{\bsnm{Perkins}, \binits{G.}},
\bauthor{\bsnm{Geary}, \binits{M.}},
\bauthor{\bsnm{Walsh}, \binits{F.}},
\bauthor{\bsnm{Nagle}, \binits{U.}},
\bauthor{\bsnm{O{\^{a}}€™malley}, \binits{L.}},
\bauthor{\bsnm{Katakam}, \binits{N.}},
\bauthor{\bsnm{White}, \binits{H.}},
\bauthor{\bsnm{Tanton}, \binits{E.}},
\bauthor{\bsnm{Hamilton}, \binits{R.}},
\bauthor{\bsnm{Glanowski}, \binits{H.}},
\bauthor{\bsnm{Forde}, \binits{E.}},
\bauthor{\bsnm{Macdonald}, \binits{C.}},
\bauthor{\bsnm{McKay}, \binits{L.}},
\bauthor{\bsnm{Edoziern}, \binits{L.}},
\bauthor{\bsnm{Doran}, \binits{P.}},
\bauthor{\bsnm{Dillon}, \binits{J.}},
\bauthor{\bsnm{Taylor}, \binits{C.}},
\bauthor{\bsnm{Evans}, \binits{P.}},
\bauthor{\bsnm{Miller}, \binits{V.}},
\bauthor{\bsnm{Wayne}, \binits{C.}},
\bauthor{\bsnm{Tebbutt}, \binits{J.}},
\bauthor{\bsnm{Hendy}, \binits{E.}},
\bauthor{\bsnm{O{\^{a}}€™brien}, \binits{P.}},
\bauthor{\bsnm{Subair}, \binits{S.}},
\bauthor{\bsnm{Dent}, \binits{H.}},
\bauthor{\bsnm{Mallet}, \binits{C.}},
\bauthor{\bsnm{Quenby}, \binits{S.}},
\bauthor{\bsnm{Hillen}, \binits{J.}},
\bauthor{\bsnm{Young}, \binits{P.}},
\bauthor{\bsnm{Harrison}, \binits{T.}},
\bauthor{\bsnm{Wood}, \binits{L.}},
\bauthor{\bsnm{Arya}, \binits{R.}},
\bauthor{\bsnm{Roughley}, \binits{L.}},
\bauthor{\bsnm{Sorinola}, \binits{O.}},
\bauthor{\bsnm{Rogers}, \binits{C.}},
\bauthor{\bsnm{Phipps}, \binits{J.}},
\bauthor{\bsnm{Arndtz}, \binits{B.}},
\bauthor{\bsnm{Azzopardi}, \binits{D.}},
\bauthor{\bsnm{Chivers}, \binits{Z.}},
\bauthor{\bsnm{Cole}, \binits{A.}},
\bauthor{\bsnm{Parmar}, \binits{M.}},
\bauthor{\bsnm{Roberts}, \binits{T.}},
\bauthor{\bsnm{Sanders}, \binits{J.}},
\bauthor{\bsnm{Tuffnell}, \binits{D.}},
\bauthor{\bsnm{Ashby}, \binits{D.}},
\bauthor{\bsnm{Norman}, \binits{J.}},
\bauthor{\bsnm{Shennan}, \binits{A.}},
\bauthor{\bsnm{Spiby}, \binits{H.}},
\bauthor{\bsnm{Tin}, \binits{W.}}:
\batitle{{Computerised interpretation of fetal heart rate during labour
  (INFANT): a randomised controlled trial}}.
\bjtitle{The Lancet}
\bvolume{389}(\bissue{10080}),
\bfpage{1719}--\blpage{1729}
(\byear{2017})
\doiurl{10.1016/S0140-6736(17)30568-8}
\end{barticle}
\endbibitem

%%% 23
\bibitem[\protect\citeauthoryear{Chud{\'{a}}{\v{c}}ek
  et~al.}{2014}]{chudacek2014open}
\begin{barticle}
\bauthor{\bsnm{Chud{\'{a}}{\v{c}}ek}, \binits{V.}},
\bauthor{\bsnm{Spilka}, \binits{J.}},
\bauthor{\bsnm{Bur{\v{s}}a}, \binits{M.}},
\bauthor{\bsnm{Janků}, \binits{P.}},
\bauthor{\bsnm{Hruban}, \binits{L.}},
\bauthor{\bsnm{Huptych}, \binits{M.}},
\bauthor{\bsnm{Lhotsk{\'{a}}}, \binits{L.}}:
\batitle{{Open access intrapartum CTG database}}.
\bjtitle{BMC Pregnancy and Childbirth}
\bvolume{14}(\bissue{1}),
\bfpage{1}--\blpage{12}
(\byear{2014})
\doiurl{10.1186/1471-2393-14-16/FIGURES/2}
\end{barticle}
\endbibitem

%%% 24
\bibitem[\protect\citeauthoryear{Georgieva
  et~al.}{2019}]{georgieva2019computer}
\begin{barticle}
\bauthor{\bsnm{Georgieva}, \binits{A.}},
\bauthor{\bsnm{Abry}, \binits{P.}},
\bauthor{\bsnm{Chud{\'{a}}{\v{c}}ek}, \binits{V.}},
\bauthor{\bsnm{Djuri{\'{c}}}, \binits{P.M.}},
\bauthor{\bsnm{Frasch}, \binits{M.G.}},
\bauthor{\bsnm{Kok}, \binits{R.}},
\bauthor{\bsnm{Lear}, \binits{C.A.}},
\bauthor{\bsnm{Lemmens}, \binits{S.N.}},
\bauthor{\bsnm{Nunes}, \binits{I.}},
\bauthor{\bsnm{Papageorghiou}, \binits{A.T.}},
\bauthor{\bsnm{Quirk}, \binits{G.J.}},
\bauthor{\bsnm{Redman}, \binits{C.W.G.}},
\bauthor{\bsnm{Schifrin}, \binits{B.}},
\bauthor{\bsnm{Spilka}, \binits{J.}},
\bauthor{\bsnm{Ugwumadu}, \binits{A.}},
\bauthor{\bsnm{Vullings}, \binits{R.}}:
\batitle{{Computer-based intrapartum fetal monitoring and beyond: A review of
  the 2nd Workshop on Signal Processing and Monitoring in Labor (October 2017,
  Oxford, UK)}}.
\bjtitle{Acta Obstetricia et Gynecologica Scandinavica}
\bvolume{98}(\bissue{9}),
\bfpage{1207}--\blpage{1217}
(\byear{2019})
\doiurl{10.1111/AOGS.13639}
\end{barticle}
\endbibitem

%%% 25
\bibitem[\protect\citeauthoryear{}{}]{uci2024ctg}
\begin{botherref}
{Cardiotocography - UCI Machine Learning Repository}.
\url{https://archive.ics.uci.edu/dataset/193/cardiotocography}
\end{botherref}
\endbibitem

%%% 26
\bibitem[\protect\citeauthoryear{Brocklehurst
  et~al.}{2017}]{brocklehurst2017infant}
\begin{barticle}
\bauthor{\bsnm{Brocklehurst}, \binits{P.}},
\bauthor{\bsnm{Field}, \binits{D.J.}},
\bauthor{\bsnm{Juszczak}, \binits{E.}},
\bauthor{\bsnm{Kenyon}, \binits{S.}},
\bauthor{\bsnm{Linsell}, \binits{L.}},
\bauthor{\bsnm{Newburn}, \binits{M.}},
\bauthor{\bsnm{Plachcinski}, \binits{R.}},
\bauthor{\bsnm{Quigley}, \binits{M.}},
\bauthor{\bsnm{Schroeder}, \binits{L.}},
\bauthor{\bsnm{Steer}, \binits{P.}}:
\batitle{{The INFANT trial}}.
\bjtitle{The Lancet}
\bvolume{390}(\bissue{10089}),
\bfpage{28}
(\byear{2017})
\doiurl{10.1016/S0140-6736(17)31594-5}
\end{barticle}
\endbibitem

%%% 27
\bibitem[\protect\citeauthoryear{Hug et~al.}{2022}]{hug2022global}
\begin{barticle}
\bauthor{\bsnm{Hug}, \binits{L.}},
\bauthor{\bsnm{You}, \binits{D.}},
\bauthor{\bsnm{Blencowe}, \binits{H.}},
\bauthor{\bsnm{Mishra}, \binits{A.}},
\bauthor{\bsnm{Wang}, \binits{Z.}},
\bauthor{\bsnm{Fix}, \binits{M.J.}},
\bauthor{\bsnm{Wakefield}, \binits{J.}},
\bauthor{\bsnm{Moran}, \binits{A.C.}},
\bauthor{\bsnm{Gaigbe-Togbe}, \binits{V.}},
\bauthor{\bsnm{Suzuki}, \binits{E.}},
\bauthor{\bsnm{Blau}, \binits{D.M.}},
\bauthor{\bsnm{Cousens}, \binits{S.}},
\bauthor{\bsnm{Creanga}, \binits{A.}},
\bauthor{\bsnm{Croft}, \binits{T.}},
\bauthor{\bsnm{Hill}, \binits{K.}},
\bauthor{\bsnm{Joseph}, \binits{K.S.}},
\bauthor{\bsnm{Maswime}, \binits{S.}},
\bauthor{\bsnm{McClure}, \binits{E.M.}},
\bauthor{\bsnm{Pattinson}, \binits{R.}},
\bauthor{\bsnm{Pedersen}, \binits{J.}},
\bauthor{\bsnm{Smith}, \binits{L.K.}},
\bauthor{\bsnm{Zeitlin}, \binits{J.}},
\bauthor{\bsnm{Alkema}, \binits{L.}}:
\batitle{{Global, Regional, and National Estimates and Trends in Stillbirths
  from 2000 to 2019: A Systematic Assessment}}.
\bjtitle{Obstetrical and Gynecological Survey}
\bvolume{77}(\bissue{2}),
\bfpage{83}--\blpage{85}
(\byear{2022})
\doiurl{10.1097/01.OGX.0000816512.11007.84}
\end{barticle}
\endbibitem

%%% 28
\bibitem[\protect\citeauthoryear{Georgieva
  et~al.}{2013}]{georgieva2013artificial}
\begin{barticle}
\bauthor{\bsnm{Georgieva}, \binits{A.}},
\bauthor{\bsnm{Payne}, \binits{S.J.}},
\bauthor{\bsnm{Moulden}, \binits{M.}},
\bauthor{\bsnm{Redman}, \binits{C.W.G.}}:
\batitle{{Artificial neural networks applied to fetal monitoring in labour}}.
\bjtitle{Neural Computing and Applications}
\bvolume{22}(\bissue{1}),
\bfpage{85}--\blpage{93}
(\byear{2013})
\doiurl{10.1007/S00521-011-0743-Y/FIGURES/6}
\end{barticle}
\endbibitem

%%% 29
\bibitem[\protect\citeauthoryear{Petrozziello
  et~al.}{2019}]{petrozziello2019multimodal}
\begin{barticle}
\bauthor{\bsnm{Petrozziello}, \binits{A.}},
\bauthor{\bsnm{Redman}, \binits{C.W.G.}},
\bauthor{\bsnm{Papageorghiou}, \binits{A.T.}},
\bauthor{\bsnm{Jordanov}, \binits{I.}},
\bauthor{\bsnm{Georgieva}, \binits{A.}}:
\batitle{{Multimodal Convolutional Neural Networks to Detect Fetal Compromise
  during Labor and Delivery}}.
\bjtitle{IEEE Access}
\bvolume{7},
\bfpage{112026}--\blpage{112036}
(\byear{2019})
\doiurl{10.1109/ACCESS.2019.2933368}
\end{barticle}
\endbibitem

%%% 30
\bibitem[\protect\citeauthoryear{Feng et~al.}{2023}]{feng2023hybrid}
\begin{barticle}
\bauthor{\bsnm{Feng}, \binits{J.}},
\bauthor{\bsnm{Liang}, \binits{J.}},
\bauthor{\bsnm{Qiang}, \binits{Z.}},
\bauthor{\bsnm{Hao}, \binits{Y.}},
\bauthor{\bsnm{Li}, \binits{X.}},
\bauthor{\bsnm{Li}, \binits{L.}},
\bauthor{\bsnm{Chen}, \binits{Q.}},
\bauthor{\bsnm{Liu}, \binits{G.}},
\bauthor{\bsnm{Wei}, \binits{H.}}:
\batitle{{A hybrid stacked ensemble and Kernel SHAP-based model for intelligent
  cardiotocography classification and interpretability}}.
\bjtitle{BMC Medical Informatics and Decision Making}
\bvolume{23}(\bissue{1}),
\bfpage{1}--\blpage{12}
(\byear{2023})
\doiurl{10.1186/S12911-023-02378-Y/TABLES/7}
\end{barticle}
\endbibitem

%%% 31
\bibitem[\protect\citeauthoryear{Spilka et~al.}{2016}]{signorini2016advanced}
\begin{barticle}
\bauthor{\bsnm{Spilka}, \binits{J.}},
\bauthor{\bsnm{Chud{\'{a}}{\v{C}}ek}, \binits{V.}},
\bauthor{\bsnm{Huptych}, \binits{M.}},
\bauthor{\bsnm{Leonarduzzi}, \binits{R.}},
\bauthor{\bsnm{Abry}, \binits{P.}},
\bauthor{\bsnm{Doret}, \binits{M.}}:
\batitle{{Intrapartum fetal heart rate classification: Cross-database
  evaluation}}.
\bjtitle{IFMBE Proceedings}
\bvolume{57},
\bfpage{1193}--\blpage{1198}
(\byear{2016})
\doiurl{10.1007/978-3-319-32703-7{\_}232/CO}
\end{barticle}
\endbibitem

%%% 32
\bibitem[\protect\citeauthoryear{Lin et~al.}{2024}]{lin2024deep}
\begin{botherref}
\oauthor{\bsnm{Lin}, \binits{Z.}},
\oauthor{\bsnm{Liu}, \binits{X.}},
\oauthor{\bsnm{Wang}, \binits{N.}},
\oauthor{\bsnm{Li}, \binits{R.}},
\oauthor{\bsnm{Liu}, \binits{Q.}},
\oauthor{\bsnm{Ma}, \binits{J.}},
\oauthor{\bsnm{Wang}, \binits{L.}},
\oauthor{\bsnm{Wang}, \binits{Y.}},
\oauthor{\bsnm{Hong}, \binits{S.}}:
{Deep Learning with Information Fusion and Model Interpretation for Health
  Monitoring of Fetus based on Long-term Prenatal Electronic Fetal Heart Rate
  Monitoring Data}
(2024)
\end{botherref}
\endbibitem

%%% 33
\bibitem[\protect\citeauthoryear{Jones et~al.}{2024}]{jones2024performance}
\begin{botherref}
\oauthor{\bsnm{Jones}, \binits{G.D.}},
\oauthor{\bsnm{Cooke}, \binits{W.R.}},
\oauthor{\bsnm{Vatish}, \binits{M.}},
\oauthor{\bsnm{Redman}, \binits{C.W.G.}},
\oauthor{\bsnm{Pan}, \binits{Y.}}:
{A Performance Evaluation of Computerised Antepartum Fetal Heart Rate
  Monitoring: The Dawes-Redman Algorithm at Term}
(2024)
\end{botherref}
\endbibitem

%%% 34
\bibitem[\protect\citeauthoryear{Romagnoli
  et~al.}{2020}]{romagnoli2020annotation}
\begin{barticle}
\bauthor{\bsnm{Romagnoli}, \binits{S.}},
\bauthor{\bsnm{Sbrollini}, \binits{A.}},
\bauthor{\bsnm{Burattini}, \binits{L.}},
\bauthor{\bsnm{Marcantoni}, \binits{I.}},
\bauthor{\bsnm{Morettini}, \binits{M.}},
\bauthor{\bsnm{Burattini}, \binits{L.}}:
\batitle{{Annotation dataset of the cardiotocographic recordings constituting
  the “CTU-CHB intra-partum CTG database”}}.
\bjtitle{Data in Brief}
\bvolume{31},
\bfpage{105690}
(\byear{2020})
\doiurl{10.1016/J.DIB.2020.105690}
\end{barticle}
\endbibitem

%%% 35
\bibitem[\protect\citeauthoryear{Sbrollini
  et~al.}{2017}]{sbrollini2017ctganalyzer}
\begin{barticle}
\bauthor{\bsnm{Sbrollini}, \binits{A.}},
\bauthor{\bsnm{Agostinelli}, \binits{A.}},
\bauthor{\bsnm{Burattini}, \binits{L.}},
\bauthor{\bsnm{Morettini}, \binits{M.}},
\bauthor{\bsnm{Di~Nardo}, \binits{F.}},
\bauthor{\bsnm{Fioretti}, \binits{S.}},
\bauthor{\bsnm{Burattini}, \binits{L.}}:
\batitle{{CTG Analyzer: A graphical user interface for cardiotocography}}.
\bjtitle{Proceedings of the Annual International Conference of the IEEE
  Engineering in Medicine and Biology Society, EMBS}
\bvolume{1},
\bfpage{2606}--\blpage{2609}
(\byear{2017})
\doiurl{10.1109/EMBC.2017.8037391}
\end{barticle}
\endbibitem

%%% 36
\bibitem[\protect\citeauthoryear{Jones et~al.}{2022}]{jones2022computerized}
\begin{barticle}
\bauthor{\bsnm{Jones}, \binits{G.D.}},
\bauthor{\bsnm{Cooke}, \binits{W.R.}},
\bauthor{\bsnm{Vatish}, \binits{M.}},
\bauthor{\bsnm{Redman}, \binits{C.W.G.}},
\bauthor{\bsnm{Pan}, \binits{Y.}}:
\batitle{{Computerized Analysis of Antepartum Cardiotocography: A Review}}.
\bjtitle{Maternal-Fetal Medicine}
\bvolume{4}(\bissue{2}),
\bfpage{130}--\blpage{140}
(\byear{2022})
\doiurl{10.1097/FM9.0000000000000141}
\end{barticle}
\endbibitem

\end{thebibliography}

\end{document}